\documentclass[journal]{IEEEtran}

\usepackage[pdftex]{graphicx}

\usepackage{amsmath}
\usepackage{amssymb}
\usepackage{multirow}
\usepackage{diagbox}
\usepackage{subfigure}
\usepackage{array}

\newcommand{\myPara}[1]{\vspace{.05in}\noindent\textbf{#1}}

\graphicspath{{./figures/}}
\usepackage{color}

\begin{document}
\title{Receptive Multi-granularity Representation for Person Re-Identification}

\author{Guanshuo Wang, Yufeng Yuan, Jiwei Li, Shiming Ge, \IEEEmembership{Senior Member,~IEEE}, Xi Zhou
\thanks{G. Wang is with the Cooperative Medianet Innovation Center, Shanghai Jiao Tong University, Shanghai, 200240, China (e-mail: guanshuo.wang@sjtu.edu.cn).}
\thanks{Y. Yuan, J. Li, X. Zhou are with Cloudwalk Technology, Shanghai, 201203, China (e-mail: \{yuanyufeng,lijiewei,zhouxi\}@cloudwalk.cn).}
\thanks{S. Ge is with the Institute of Information Engineering, Chinese Academy of Sciences, Beijing, 100095, China. }
\thanks{S. Ge is the corresponding author. (e-mail: geshiming@iie.ac.cn)} 
}

%
%

\markboth{IEEE Transactions on Image Processing}%
{Shell \MakeLowercase{\textit{et al.}}: Bare Demo of IEEEtran.cls for IEEE Journals}


\maketitle

\begin{abstract}
A key for person re-identification is achieving consistent local details for discriminative representation across variable environments. Current stripe-based feature learning approaches have delivered impressive accuracy, but do not make a proper trade-off between diversity, locality, and robustness, which easily suffers from part semantic inconsistency for the conflict between rigid partition and misalignment. This paper proposes a receptive multi-granularity learning approach to facilitate stripe-based feature learning. This approach performs local partition on the intermediate representations to operate receptive region ranges, rather than current approaches on input images or output features, thus can enhance the representation of locality while remaining proper local association. Toward this end, the local partitions are adaptively pooled by using significance-balanced activations for uniform stripes. Random shifting augmentation is further introduced for a higher variance of person appearing regions within bounding boxes to ease misalignment. By two-branch network architecture, different scales of discriminative identity representation can be learned. In this way, our model can provide a more comprehensive and efficient feature representation without larger model storage costs. Extensive experiments on intra-dataset and cross-dataset evaluations demonstrate the effectiveness of the proposed approach. Especially, our approach achieves a state-of-the-art accuracy of 96.2\%@Rank-1 or 90.0\%@mAP on the challenging Market-1501 benchmark.
\end{abstract}


%
\IEEEpeerreviewmaketitle

\section{Introduction}
\IEEEPARstart{T}{he} task of person re-identification (re-ID) aims to retrieve images of a specific person among cross-camera gallery pedestrian databases captured in the wild. Its main challenge arises from the large appearance variations on pose or clothes, heavy occlusions, background clutter, and detection failure. Thus, it is necessary to find a pivotal solution to address such a captured-in-wild challenge: how to perform robust identity-discriminative representation for facilitating person re-ID under unconstrained scenarios. General image retrieval methods such as LRGA \cite{5989829}, focusing on data ranking by learning robust functions and representation refinement via statistical approaches. Due to the dependency on ranking according to distance or similarity functions between low-dimension features, person re-ID is also a query-dependent common image retrieval problem, solvable by these methods. However, general methods do not consider some necessary prior knowledge, \textit{e.g.} body parts, body joints, vertical feature distributions, while existing re-ID targeted methods apply as much as such knowledge to feature representations to supplement hard information for naive feature learning. 

\begin{figure*}
	\includegraphics[width=1.0\linewidth]{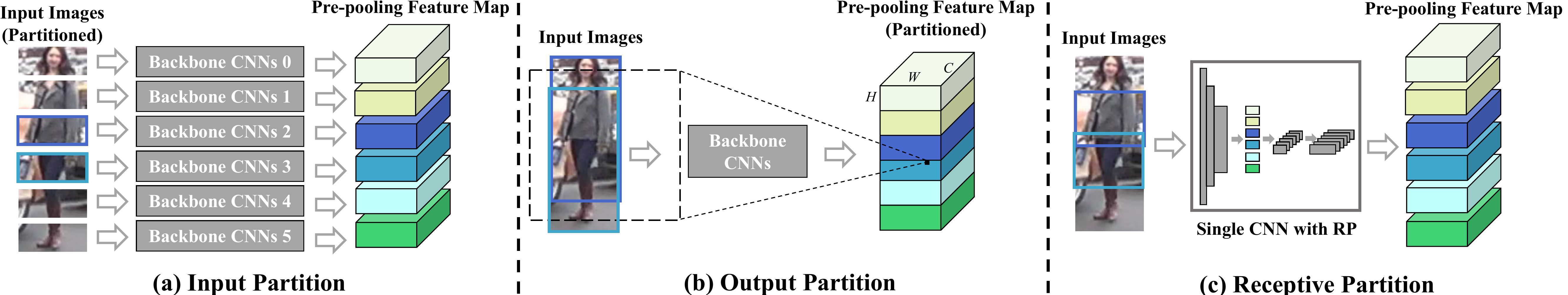}
	\centering
	\caption{\label{fig1}Comparison of stripe-based feature learning for person re-ID with output (a), input (b) and our proposed receptive partition (c) operations for local feature representations. For all the methods, output pre-pooling feature maps are formed into several stripe partitions corresponding to local part regions, and then globally pooled and reduced into components of final feature vectors. Different partition methods bring various ranges of receptive regions for local feature stripes ($e.g.$ the 3th and 4th stripe). Especially, the most commonly applied output partition actually represent a relatively global regions instead of just the corresponding local stripes, which comes from the large receptive field for each neuron on the final feature map.}
\end{figure*}

Due to the powerful representation ability, solutions based on Convolutional Neural Networks (CNNs) have become the mainstream and achieved remarkable performance. Especially, recent local representation approaches \cite{cheng2016person,li2018harmoniou,sun2018beyond,zhang2017alignedreid,su2017pose,zhao2017spindle,zhao2017deeply,li2017learning,liu2017end,liu2017hydraplus,borgia2018cross,feng2018learning,zheng2019pose} which extract features from local body parts by CNNs, have strongly proven robust to the environment variations. Among all these local representation methods, stripe-based feature learning \cite{yi2014deep,cheng2016person,sun2018beyond,wang2018mgn,zheng2018coarse,huang2018eanet,sun2019perceive,fu2019horizontal,yao2017deep} achieves the best trade-off between performance and efficiency. Fig. \ref{fig1} (a) and (b) show two standard stripe partition schemes. For input partition (a), the input images are split into several uniform horizontal stripe patches, and respectively fed into independent backbone CNNs. Then outputs feature stripes before pooling to represent each local body parts. Recent popular output partition (b) methods are more computational friendly and simpler in practical application. The input pedestrian image is fed into a deep backbone network, then the whole last feature map before the global pooling operation is split into several horizontal feature stripes. For both kinds of methods, output feature stripes are globally pooled in both spatial dimensions and reduced by 1x1 convolutions in the channel dimension, then outputs as a local feature fed into independent classification or distance learning tasks. Stripes are considered as local representations corresponding to some areas in the input pedestrian image. At last, all the feature vectors are concatenated in the channel dimension as the final representation for a pedestrian image. In a word, the core idea of stripe-based feature learning is to resolve the whole learning task to several sub-tasks on local partitions, coarsely determined by uniform horizontal stripes split on the final output feature map in the deep network.

Between these two partition methods, another difference that might be easily overlooked is the equivalent receptive region areas of local partitions. Considering each neuron in output feature maps has its corresponding input image area or its receptive field, each feature stripe also corresponds to some adjacent area covered by fields. For commonly used ultra-deep network backbone architecture (e.g., ResNet-50), as shown in Fig. \ref{fig1} (b), the sizeable receptive field of neurons on final output feature maps can cover the whole normal-sized input images or local patches. Based on this setting, the actual receptive region for each feature stripe changes with different partition methods. For input partition directly on pedestrian images, receptive regions are restricted within each input image patches so that locality can be maximized. However, rigid stripes without any association between each other intensify the effects of body part misalignment that commonly happens in person re-ID. Instead, output partition for uniform stripes on the final feature maps introduces locality not as complete as we wish \cite{sun2018beyond,sun2019perceive}, which distracts local attention and brings redundancy between local features. 

Besides, stripe-based feature learning methods are also limited by the rigid partition. On the one hand, local part information is not ideally arranged uniformly along the height dimension. Current uniform partition strategies do not match the actual part semantic distribution, which may cause consistency issues to affect the representation efficiency. On the other hand, due to the fixed numbers of uniformly split feature stripes, the representation granularity is restricted in a fixed state. However, part local information is not likely to represent only a single scale. Features with large (e.g., clothes color), middle (e.g., accessory), or small (e.g., cloth logo) granularities may all express critical identity information. Not to mention that the body misalignment within actual pedestrian images can intensify this situation.

In this paper, to address these problems, we propose receptive multi-granularity learning (RMGL) approach for feature representation in person re-ID task. As the core of stripe-based learning, partition strategy determines the form and context of discriminative features. Towards the problems in input and output partition, we introduce the receptive partition to ensure relevance while enhancing locality. As shown in Fig. \ref{fig1} (c), the actual receptive image region for each neuron before global pooling is restricted in a much smaller range by partitioning on intermediate feature maps in a proper level of backbone networks, and all the split stripes are fed into the subsequent network without extra model parameters. Towards the metric of partition, we devise activation balanced pooling strategy to determine the coarse region of each stripe on the final feature map in testing phases, instead of uniform approaches. The numbers of selected global maximum activations in all the split stripe feature regions are ensured balanced. To relieve the challenging misalignment issues, we propose random shifting augmentation employed in training phases. By randomly sampling the major pedestrian bodies and re-locating it by cropping, the distribution of regions within the bounding boxes where persons appear can be disturbed for better robustness to misalignment. Considering the unique paradigm of stripe-based representations and multi-granularity learning \cite{chen2011short,ni2014multiple,wang2015multiple,yao2018multiple} that comprehensively complements the diversity of appearance information from inputs, we develop a multi-branch network architecture to introduce multiple granularities in feature representations, where stripe features are respectively partitioned into different numbers to control the scales.

Our main contributions are as follows. First, we propose a receptive multi-granularity learning approach to comprehensively represent local features from different scales of multi-granularity for high-accuracy person re-ID. Different from existing approaches, we introduce receptive paths by a novel partition strategy operated on the receptive maps. Besides, we combine the advantages of RP and OP for mutual performance improvement and introduce the dual-path architecture with an original path and a receptive partitioned path by shared kernel weights, which explores the potential locality representation without more model parameters. Second, we develop an adaptive partition pooling method according to activation significance on feature maps, which alternates the semantic-level spatial balance assumption to feature-level activation significance balance assumption. We also introduce a simple yet effective data augmentation scheme to shift the location of pedestrians in images to enhance the robustness of misalignment. Finally, we conduct comprehensive experiments on several mainstream datasets \cite{zheng2015scalable, zheng2017unlabeled, li2014deepreid, wei2018person} to verify the effectiveness of our proposed approach.

\section{Related Works}
\textbf{Deep Learning for Person re-ID.} The evolution from hand-crafted feature descriptors to representing features by deep learning pushes the performance of person re-ID methods to an unprecedented level. Mainstream person re-ID by deep learning paradigm can mainly divided into supervised and unsupervised methods. We evaluate both intra-dataset and cross-dataset settings in our experiments, so here we introduce both kinds of methods in this section.

For supervised learning, \cite{li2014deepreid,yi2014deep} first introduces a deep convolutional network to the person re-ID field in the early era, which both choose Siamese-like network to learn the matching relationship between representations of pedestrian image pairs, especially \cite{yi2014deep} first noticed the importance of local feature learning in deep re-ID. \cite{ahmed2015improved,varior2016gated} focuses on utilizing the relationship of intermediate feature maps to improve the discrimination of representations. \cite{lin2019improving} introduces auxiliary pedestrian attributes to improve identity representation learning. When realizing the limitation of purely global feature learning, many attempts to local feature learning haven arisen.  Some methods \cite{zhao2017spindle,su2017pose,suh2018part,huang2018eanet,zhang2018densely} refer to external clues of pose estimation or body part parsing to extract body part features of persons. \cite{zhao2017spindle,su2017pose} utilize the structural part by pose estimation prediction to form relatively precise local region proposals for further representations. \cite{suh2018part} implements bilinear pooling between appearance features and body key point estimation features for part alignment on identity features. \cite{huang2018eanet} stands on basic stripe-based feature learning and introduces structural pose clues for proper stripe splitting. \cite{zhang2018densely} introduces 3D human body parsing to establish a semantic association between appearance and parsing results. \cite{li2017learning,zhao2017deeply,yao2017deep,li2018harmoniou} extract local features from self-proposal local part regions, coincidentally all these methods choose Spatial Transformer module \cite{jaderberg2015spatial} for region proposal by self-learning. \cite{liu2017end,liu2017hydraplus,li2018harmoniou,wang2018mancs} introduce a multi-level soft attention mechanism to enhance discrimination on part detail information of representations in spatial or channel dimensions. Stripe-based learning methods \cite{sun2018beyond,wang2018mgn,fu2019horizontal,zheng2018coarse,huang2018eanet} commonly split the last feature map into several stripes as local representations on human bodies. \cite{cheng2016person,hermans2017defense,chen2017beyond,zheng2018coarse} rely on triplet or more complicated losses for distance metric learning. \cite{zheng2017unlabeled,zhong2018camera,liu2018pose} utilize Generative Adversarial Network (GAN) \cite{goodfellow2014generative} to augment image samples with different distributions. \cite{shen2018deep,shen2018person,luo2018spectral} mine the graph relationship between sample pairs for person re-ID. \cite{zheng2019joint} combines feature learning and image generation to improve the re-ID feature discrimination. \cite{quan2019auto} first introduces an effective AutoML framework into deep person re-ID, which particularly targets retrieval tasks.

Notice that identity annotations for large scale pedestrian images across cameras used for training from scratch or finetuning on pretrained models with supervised learning are very expensive to obtain. Unsupervised person re-ID methods are significant for better-generalized pedestrian retrieval. Peng \textit{et al.}  \cite{peng2016unsupervised} uses a multi-task dictionary learning strategy to learn domain-invariant but ID-discriminative representations with unlabeled target datasets. Yu \textit{et al.} \cite{yu2017cross} propose a model to learn an asymmetric metric for shared view-invariance feature spaces. Wang \textit{et al.} \cite{wang2018transferable} aims to learn an attribute and identity combined feature representation transferable to new domains. Lv \textit{et al.} \cite{lv2018unsupervised} introduce Spatio-temporal patterns learned from unlabeled target datasets with available temporal information to re-rank the ranking results on target domains. Fan \textit{et al.} \cite{fan2018unsupervised} utilize self-paced learning to estimate pseudo-labels by clustering to finetune on the source domain models. Li \textit{et al.} \cite{li2018unsupervised} propose an end-to-end learning method to establish the associations between tracklets across cameras. Ding \textit{et al.} \cite{ding2019adaptive} propose an adaptive selection method to mine example annotation information and introduce balancing strategy for wider example exploration. Wu \textit{et al.} \cite{wu2019progressive} propose a progressive semi-supervised learning strategy that focuses on one-shot person re-ID tasks, \textit{i.e.} learning on the dataset including identities with only one labeled example along with many unlabeled examples. Fu \textit{et al.} \cite{fu2019self} focus on local-based unsupervised feature learning, and inherit the semi-supervised setting with self-grouping strategies.

\textbf{Feature Pooling.} To reduce the feature representation to an acceptable size, some abstracting operations, or feature ``pooling", are usually conducted in feature representations. \cite{zhou2010image} utilizes a kernel-based with spatial pyramid matching for local descriptors. \cite{xie2015task} integrates pooling operations into classification tasks. \cite{Gong2014Multi} improves the geometric invariance of CNN activations by introducing multi-scale disordered pooling. \cite{8382272} proposes a trainable pooling layer that generalizes max and average pooling. Global pooling is a simple and efficient operation \cite{lin2013network} to comprehensively merge information across the spatial domain. Especially in person re-ID, Global Max Pooling has been proved more powerful in many practices \cite{wang2018mgn,dai2018batch,zhang2019learning} by selecting significant activation. However, for tasks determined by fine-grained information, global pooling on the whole output feature maps might eliminate useful detailed clues in spatial dimensions. Existing stripe-based methods for person re-ID decomposes global pooling on the whole spatial domain into several ``local" pooling operations on uniform stripes, and respectively employs independent learning tasks in the subsequent training process. However, for existing stripe-based learning methods, the commonly used uniform partition operations quite rely on the semantic body part alignment. We propose activation balanced pooling to divide local representations with equal importance, which brings considerate improvement on feature discrimination.

\textbf{Data Augmentation for Person re-ID.} Due to conflicts between the limited scale of training sets and large variations of samples, data augmentation is an efficient way to relieve this dilemma. \cite{zhong2017random,fu2019horizontal} erase randomly-sampled regions on the input images to augment occluded image samples for robustness enhancement. \cite{dai2018batch} filters output feature maps with random masks uniformly in the same mini-batch. \cite{zhong2018camera,liu2018pose} generates samples in different capturing conditions (\textit{e.g.} cameras and poses) to augment the appearing camera of each identity to the whole candidates. We introduce the random shifting augmentation into person re-ID to disturb the fixed distribution of person appearing regions for better generalization.

\begin{figure*}
	\includegraphics[width=1.0\linewidth]{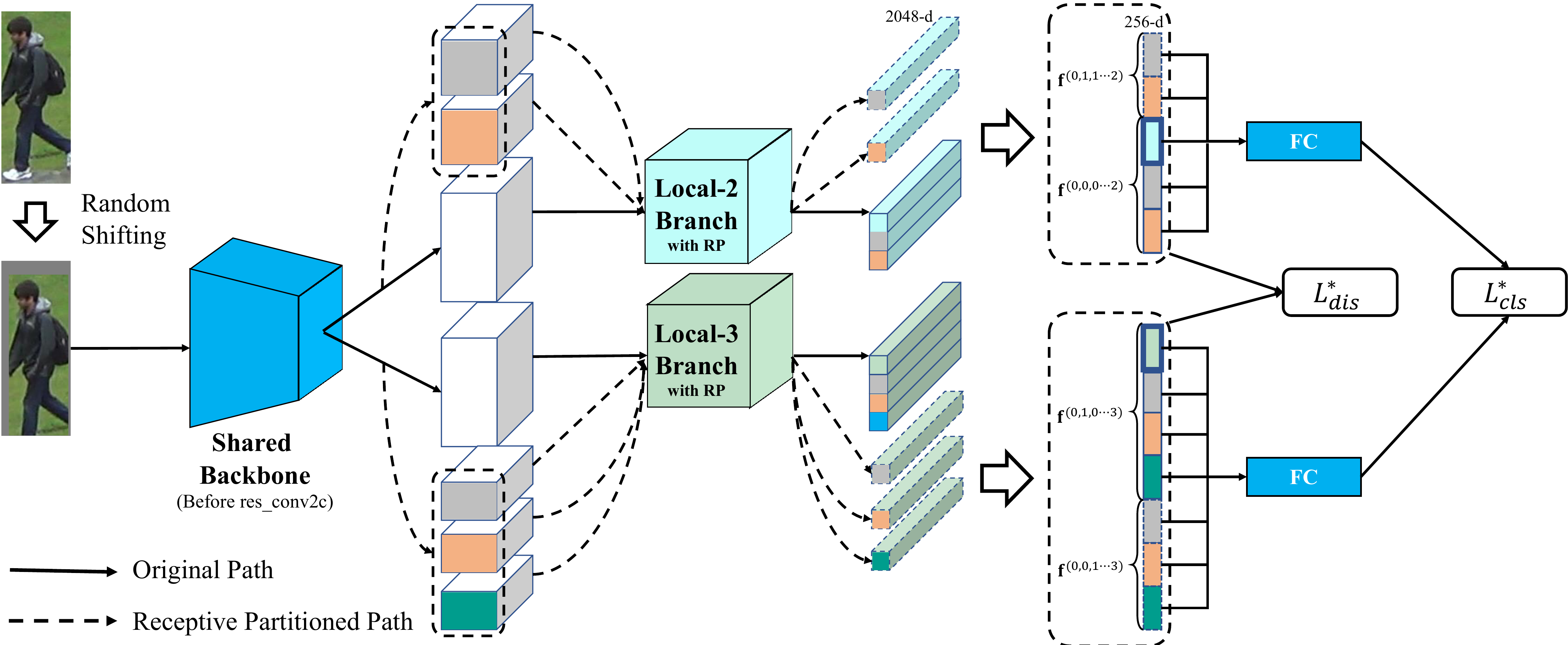}
	\centering
	\caption{\label{fig2}The RMGL framework. The input pedestrian image is first randomly augmented and then goes forward through the shared backbone network. After that, the shared backbone network is split into Local-2 and Local-3 branches with different output partitioned local feature stripes and accompanied by a receptive partitioned path for locality enhancement. Then the output stripe feature maps from both paths in two branches are globally pooled into 1-dimensional vectors. Each representation is learned for an independent classification task while the concatenation features for each branch are used to learn a distance learning task. }
\end{figure*}

\section{Methodology}

\subsection{Overview}
The proposed RMGL (see Fig. \ref{fig2}) is used to explore more potential of current local feature learning methods. Its core idea is to represent identity features with enhanced diversity, locality, and robustness. For input pedestrian image $X$, after inference by backbone networks with shared weights, the single pathway splits into multiple independent sub-branches at a proper intermediate layer of output $\mathbf{x}_m$, with the $i$-th branch representation function denoted as $g_i$. In different branches, the final output feature maps $g_i(\mathbf{x}_m)\in R^{c\times h\times w}$ are split into different numbers of partitions $K_i$. By controlling $K_i$ of different branches, various levels of pedestrian image reception locality, \textit{i.e.} granularities, can be introduced to represent person identities comprehensively. Here we take the two-branch (or bident) case as an example. $K_i$ are respectively set to 2 and 3, denoted as Local-2 and -3 Branches. Notice that a representation overlapping exists between Local-2 and -3 Branches, which mutually complements the association between local representations. Besides, all the Local Branches also represents globally by the whole output feature maps for the necessary global view reception, denoted as $x_i^{0}$.

However, just in this multi-granularity framework based on output partitioned stripe-based feature learning, potential issues except diversity are not resolved totally. Towards locality, we introduce a new receptive partition strategy for local representation and take the path represented with RP as locality guidance for both branches. Features from original path $f_{i}^{(0, k)}$ and receptive partitioned path $f_{i}^{(1, k)}$ (both $k=1...K_i$) in $i$-th branch are both represented for global and local representations. So the final features of RMGL are the concatenation of all the global and local reduced 256-d features in dual paths for both branches, representing as
\begin{equation}
\mathbf{f}=[[\mathbf{f}_{i}^{(k)}]|_{k=0}^{K_0}, [\mathbf{f}_{i}^{(k)}]|_{k=0}^{K_1}, [\mathbf{f}_{i}^{(k)}]|_{k=0}^{K_2}],
\end{equation}
where the $[\cdot]$ operation is concatenation along the channel dimension. On local feature stripes, activation balanced pooling are applied by dividing uniform activation significance, instead of traditional stripe-based features. For better generalization and robustness to misalignment, we introduce random shifting augmentation on input images during training.

All the representations are trained with classification tasks. The global feature maps and local stripes from all the paths in both branches are pooled by Global Max Pooling, and reduced into 256-d feature vectors $\mathbf{f}_{i}^{(k)}$ in numbers of channels by 1x1 convolution layers with BN layers \cite{ioffe2015batch}. Independent feature representations for both branches are trained independently by multi-task learning with classification and distance learning.

\textbf{Training loss functions} All the 256-d global or local features from different paths are respectively fed in independent fully connected layers as classification tasks, constrained by cross-entropy loss $L_{cls}$ with softmax function as
\begin{equation}
L_{cls}^{(b, p, s)}=-\sum_{i=1}^N{\log{\frac{e^{s_{i, y_i}^{(b, p, s)}}}{\sum_{k=1}^{C}{e^{s_{i, k}^{(b, p, s)}}}}}},
\end{equation}
where $N$ and $C$ respectively denote the batch size of the input mini-batch and the number of total identities in the training sets. $s_{i, k}$ denotes the predicted scores towards identity $k$. $(b,p,s)$ represents the variable corresponds to the $b$-th branch, $p$-th path and $s$-th stride in RMGL. Here $s=0$ is denoted as the global feature pooled from the whole feature maps. Metric learning tasks are also introduced, constrained by batch hard triplet loss \cite{hermans2017defense}. The loss is formulated as
\begin{equation}
\begin{aligned}
L_{dis}^{(b)}=-\sum_{i=1}^P\sum_{a=1}^K [\alpha&+\max_{\substack{p=1...K \\ p \neq a}}\|\mathbf{f}_{i, a}^{(b)}-\mathbf{f}_{i,p}^{(b)}\|_2\\&-\min_{\substack{n=1...K \\ j=1...P \\ j \neq i}}\|\mathbf{f}_{i,a}^{(b)}-\mathbf{f}_{j,n}^{(b)}\|_2]_+,
\end{aligned}
\end{equation}
where the feature vector extracted on the $a$-th image sample of identity $i$ in branch $b$ is denoted as $\mathbf{f}_{i,a}^{(b)}$. Minimizing the total training object function can be summarized as
\begin{equation}
\min_{\mathbf{w}}{L}=\sum_{b=1}^{N_b}{\left[L_{dis}^{(b)}+\sum_{p=1}^{N_p}{\sum_{s=0}^{n_s^{(p)}}{L_{cls}^{(b,p,s)}}}\right]}.
\end{equation}

In the following section, each of the components for discriminative receptive locality enhancement will be explained.

\begin{figure}
	\centering
	\subfigure[\label{fig3a} Receptive partition]{
		\centering
		\includegraphics[width=1.0\linewidth]{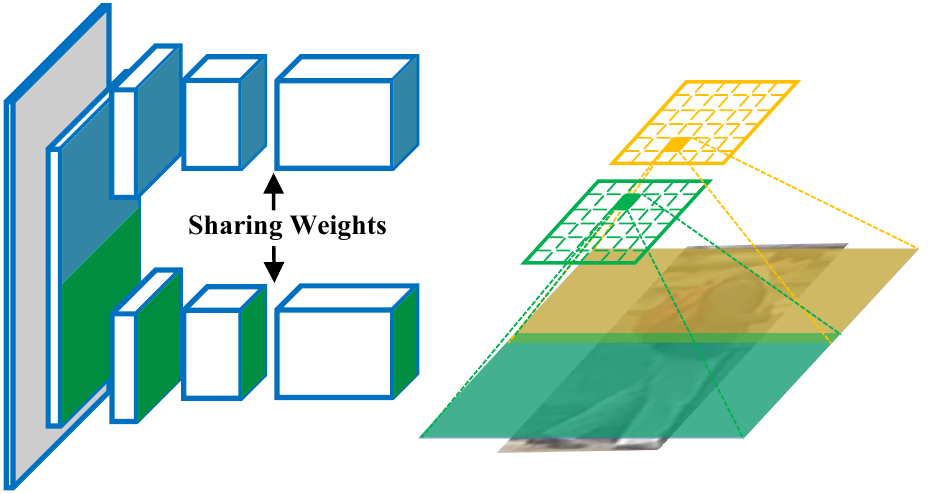}
	}
	
	\subfigure[\label{fig3b} Dual paths with receptive partitioned path]{
		\centering
		\includegraphics[width=1.0\linewidth]{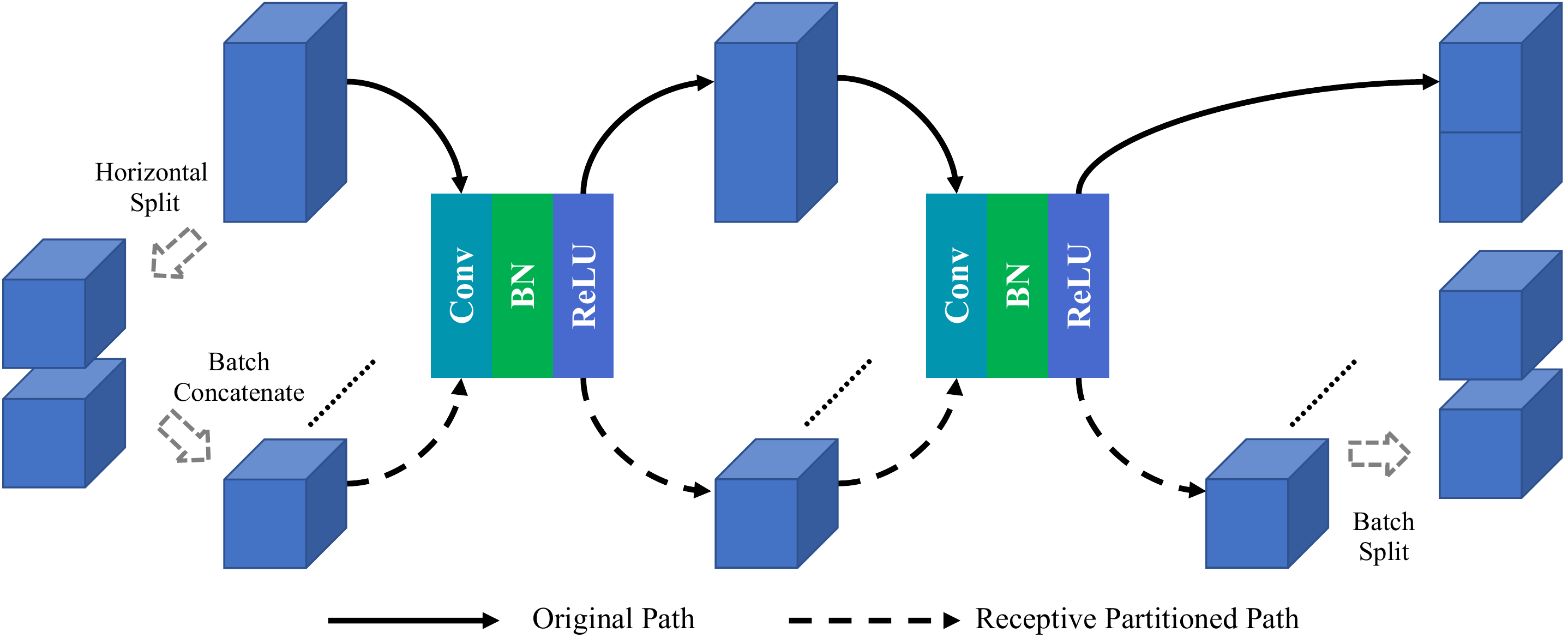}
	}
	\caption{Illustrations of receptive regions for stripe-based feature learning partitions. Assume that the input image is with the size of $384\times128$, and the backbone network is ResNet-50 with the receptive field size of 363. (a) In receptive partition, receptive regions are restricted in smaller areas with the moderate association. (b) Dual-path Conv-BN-ReLU network with the receptive partitioned path. Both paths share the same network weights. }
\end{figure}

\subsection{Receptive partition}
The motivation of stripe-based feature learning is to enrich the feature locality. However, neither output nor input partition strategy to extract local representation based on very deep network backbones \cite{simonyan2014very,szegedy2015going,he2016deep,huang2017densely} can ideally represent discriminative information in proper part region areas. In very deep architectures, due to the complex connectivity stacked from network layers, the equivalent input region of the image can be expanded to a fairly large area. For layers in CNNs, the size of receptive field from top $i$-th bottom input map to $j$-th top output map can be recursively calculated as \footnote{Effects of dilation factor are ignored for not a common setting in our task.}
\begin{equation}
\begin{aligned}
\label{rf_cal}
r_{i,j+1} &= r_{i,j}+(k_{j,j+1}-1)*\prod_{k=i}^{j}{s_{k,k+1}}, \\
&s.t. \quad i\ge0, \quad i<=j,
\end{aligned}
\end{equation}
where $s_{j,j+1}$ is the stride factor for $j$-th layer, and $k_{j,j+1}$ is the kernel size for the layer from $j$ to $j+1$. Even considering Effective Receptive Field (ERF) \cite{luo2016understanding}, the size of which distributed in Gaussian with linear growth rate $O(\sqrt{N})$ for $N$-layer deep network, it still can cover a considerable range of the input image. Here we take example with the backbone model, $i.e.$ modified ResNet-50 \cite{sun2018beyond} used in person re-ID tasks. According to Eq. \ref{rf_cal}, the receptive field of the final output neurons of stripe local features is 363 before global pooling, which means most of the contents in normal-sized pedestrian images can be undoubtedly recognized just by this neuron, and even larger after global pooling. Output partition in modern stripe-based methods does not significantly change the perception range. On the one hand, the shallower network can introduce a smaller reception field size, but sacrifices powerful representation capability. On the other hand, input images can be surely resized into a larger size comparable to the receptive field size to realize relative locality for improved accuracy performance, but increases fairly high computational costs. Besides, as mentioned in Sec. I, input partition with maximized locality at some granularity scale, but basically ignores within-part semantic consistency with inner partitions.

Towards the trade-off between locality and consistency, we introduce the receptive partition (RP) strategy into stripe-based local feature learning, illustrated in Fig. \ref{fig3a}. Different from traditional input or output partition at some granularity scale, receptive region sizes for local representations of which determined by inherent properties (\textit{e.g.} network architecture), RP controls the effective receptive areas by the partition position, a free parameter that can be flexibly operated. The $i$-th intermediate convolutional layer in the network with the output feature map of height $h_i$ is chosen, and the receptive field of this layer can be computed as Eq. \ref{rf_cal} as $r_{0, i}$. Here the intermediate feature map is divided into $n_s$ independent tensors along the $h$-dimension and feeds them into the subsequent network. Towards this partition stripe at layer $j$, this operation restricts the maximum area of the equivalent receptive local region of subsequent network layers into $\mathbf{R}_j^p$, formulated as
\begin{equation}
\label{rf_restricted}
\mathbf{R}^p_j = r_{i,j}+(\frac{h_j}{n_s}-1)*\prod_{k=0}^{j-1}{s_{k,k+1}}, \\
\end{equation}
After RP, all independent split tensors are concatenated along the dimension of batch, \textit{i.e.} with a feature map shape $n_s n \times c \times \frac{h}{n_s} \times w$. Then reshaped feature maps are fed into the subsequent network, equivalent to all the stripes are inferred by network architecture with sharing weights. Compared with traditional partitions, RP introduces no extra computational costs, but helps remove an amount of complex connectivity to extra area from the expected region corresponding to the feature stripe, which enhances locality and reduces the over-fitting risk.

\textbf{Discussion} Notice that indeed, this operation does not change the size of reception fields, which are only determined by the kernel size and stride factor. However, when the receptive field of the subsequent network $r_{j,N}$ satisfies 

\begin{equation}
\label{rf_condition}
r_{j, N}\ge \frac{h_j}{n_s},
\end{equation}

RP will cut down the spatial connectivity by the subsequent network to the outer regions of the corresponding area. The actual receptive region of each neuron at the last output layer will be restricted only in the range of $R_p$, regardless of the depth effect for the larger receptive field. Although the restricted regions by RP still involves in the receptive regions by global or OP representations, but the shielding effects on most non-corresponding regions make the local representations can concentrate more on partial details, free of global trade-off to ignore local information. This is also the reason that stripe-based methods can perform better than global, and our proposed RP carries it forward. Meanwhile, considering the Batch Normalization commonly used in backbones, RP actually does not change the inference in training or inferring. 

Empirical studies have verified the effectiveness of global representation \cite{wang2018mgn}. It still provides specific and instructive identity representation. So in our partition design, the original path is reserved, and  a new receptive partitioned path is inserted after a chosen feature map at layer $j$ with local representations by RP. The dual-path architecture is shown in Fig. \ref{fig3b}. These two network paths share the same architectures and weights in the corresponding layer, which makes the spatial size of the final output feature maps from RP the same as each stripe partition in the original path. Output feature maps are split back into $n_s$ shards along the batch dimension.

\begin{figure}
	\includegraphics[width=1.0\linewidth]{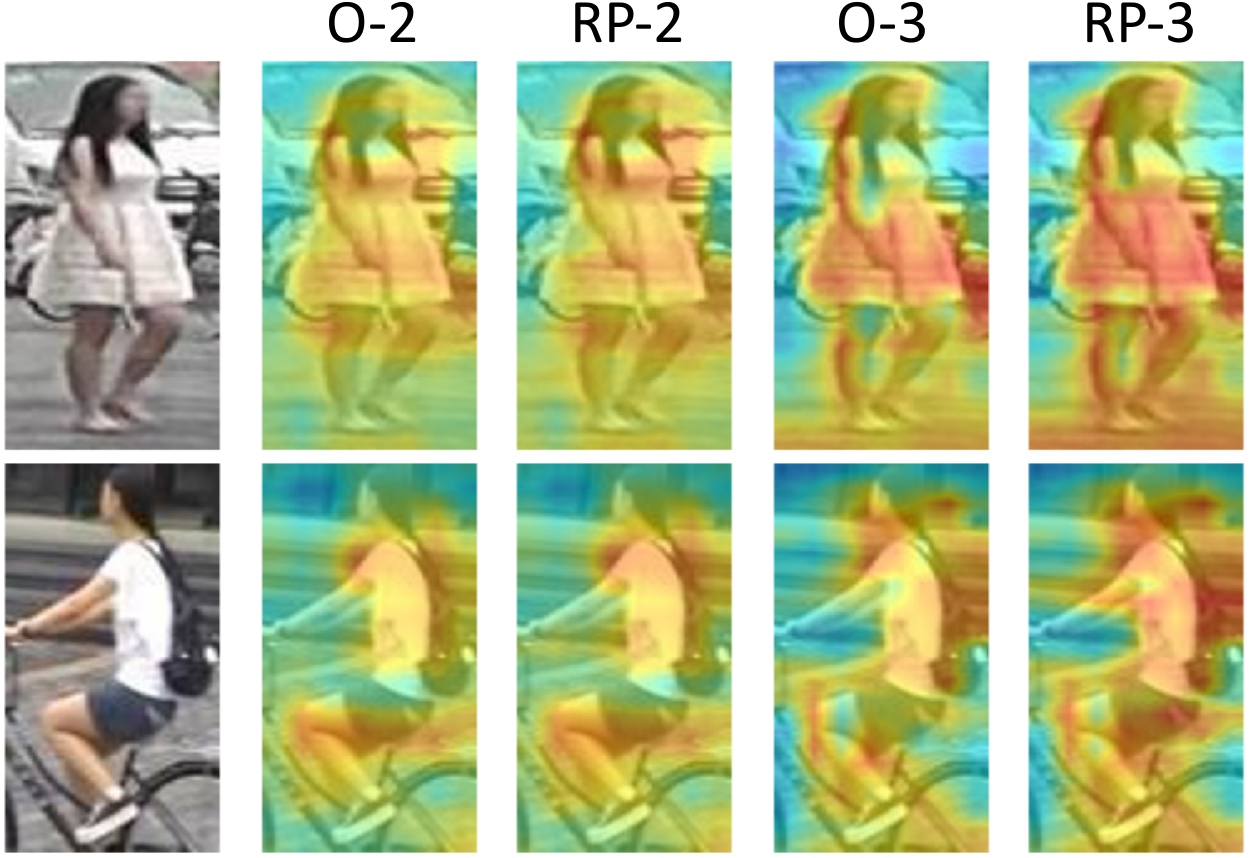}
	\centering
	\caption{\label{fig4}Visualization of responses on the last feature map by the original and receptive partitioned paths from both branches with different granularities. O-2/3 stands for the output by the original path with 2 or 3 stripes. RP-2/3 stands for the output by the receptive partitioned path with 2 or 3 stripes. }
\end{figure}

In our method, RP does not completely replace the original network path, although it is acceptable as the partition strategy for independent local representation. On the one hand, the performance improvement by stripe-based methods proves the effectiveness of stripe pooling compared to the naive global feature learning, although not contributed by so-called enhanced locality. On the other hand, RP shrinks the overlapping area between receptive regions of different stripe features to increase the requirement for within-part consistency of RP representations. The original network path can be regarded as a global representation clue to avoid extreme misalignment cases. Due to the dual-path architecture by sharing weights in each branch, better accuracy can be achieved with lighter model as 2-branch MGN. However, the intermediate feature map memory and computation costs are equal to the burden as 4-branch MGN. Fig. \ref{fig4} visualizes the effects of this dual-path architecture from different granularities. From the view of granularity, we can find that responses are more concentrated on local parts on representation with higher granularity. Comparing the original with the receptive partitioned path, the approximate response tendencies are similar, but the representation from the receptive partitioned path focuses on wider contents of pedestrian body parts.

\subsection{Activation balanced pooling}
In existing stripe-based feature learning for person re-ID, the final output feature map is split into several uniform stripes and respectively reduced into features by global pooling operations. However, this might not be a perfect solution for local representations due to some problems such as misalignment, which comes from the conflict between the assumption of within-part consistency and the infeasibility of perfect alignment of pedestrian images. Without external semantic guidance \cite{suh2018part,huang2018eanet}, considering this principle and non-locality, feature representation is neither necessary nor reasonable. Here we explore another approach of dividing partitions from the view of feature selection.

\begin{figure}
	\centering
	\includegraphics[width=1.0\linewidth]{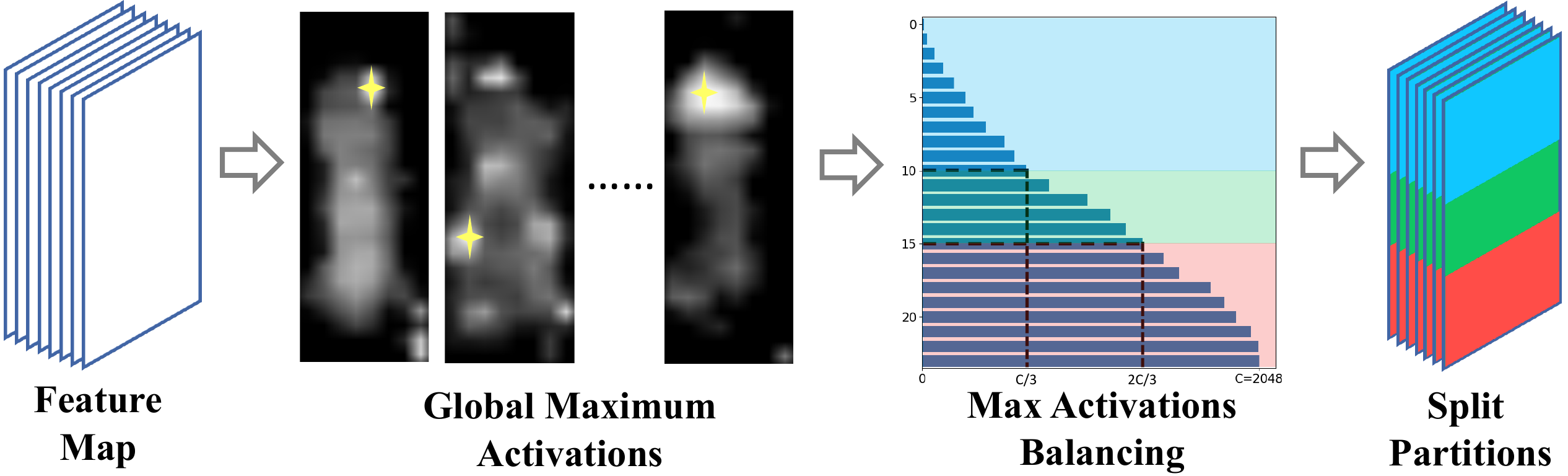}
	\caption{\label{fig5} Illustration of activation balanced pooling. Maximum activations in every channel of the pooled feature map are located. The cross symbol in Max Activation Selection stage represents the location of max values in the current channel. According to these maximum activation locations, the max activation histogram along the height dimension is computed. Then the pooled feature map is divided \textit{s.t.} balanced significance in every local feature stripe.  }
\end{figure}

Since the within-part alignment dependency cannot be easily satisfied, we turn to reconstruct a new hypothesis: assume that all the local stripe parts share equivalent importance. Based on this assumption, although the semantics of each local part cannot be definitely consistent, the efficiency of total local sub-representation learning will be comprehensively maximized. We propose the activation balanced pooling (ABP) to substitute uniform partitions in stripe-based methods.

The pipeline of ABP is shown as Fig. \ref{fig5}. For each channel in a feature map, the most significant activation is the position with maximum responses on both spatial dimensions height and width. Borrowing the idea of feature selection in spatial global max pooling, the number of maximum response occurrences are counted at each spatial position across channels. For the feature map $\mathbf{x}$ with shape $H\times W \times C$, the activation value at the location $(h, w)$ of the channel $c$ is denoted as $x_{h,w,c}$. An indicator function $I_{\mathbf{x}}(h,w,c)$ is defined to represent whether the neuron in a feature map channel is the maximum response in spatial dimensions, formulated as

\begin{equation}
I_{\mathbf{x}}(h,w,c)=\left\{
\begin{aligned}
&1, & \forall x_{i,j,c|i\neq h, j\neq w} < x_{h,w,c},\\
&0, & \exists x_{i,j,c|i\neq h, j\neq w} > x_{h,w,c},\\
\end{aligned}
\right.
\end{equation}

To further describe the relationship between the position in the vertical orientation and the accumulated number of activations, the max activation histogram function $H_{\mathbf{x}}(h)$ utilizing the indicator is calculated as follows:
\begin{equation}
\begin{aligned}
H_{\mathbf{x}}(h)=\sum_{i=1}^{h}{\sum_{j=1}^{W}{\sum_{k=1}^{C}{I_{\mathbf{x}}(i, j, k)}}},
\end{aligned}
\end{equation}
Max activation histogram makes it not difficult to describe the significance of local stripes. Here the feature map is split into $n_s$ horizontal stripes at vertical positions $[h_k]_{k=0}^{n_s}$ where each max activation histogram function $H_{\mathbf{x}}(h_k)$ value satisfies
\begin{equation}
\left\{
\begin{aligned}
&H_{\mathbf{x}}(h_k) \leq \lceil\frac{c}{C}n_s\rceil, \\
&H_{\mathbf{x}}(h_k+1) > \lceil\frac{c}{C}n_s\rceil, \\
\end{aligned}
\right.
, \forall c = 1, \cdots, n_s-1,
\end{equation}
where the bottom and top are respectively $h_0=0$ and $h_{n_s}=H$. This condition means each local stripe may hold balanced activation importance on the feature maps. After that, the partitioned feature stripe are pooled in global average as
\begin{equation}
f_k = \frac{1}{(h_{k+1}-h_k)\times W}\sum_{i=0}^{H-1}{\sum_{j=0}^{W-1}{x_{i,j,c}}},
\end{equation}
ABP aims to make all the stripe features represent identity information with balanced significance after splitting, which reduces the feature redundancy caused by misalignment. 

\textbf{Comparison with Refined Part Pooling (RPP).~} The idea of calibrating the local representation with re-pooling strategy ABP is indeed similar to the learning-based rearrangement with RPP, and both attempt to overcome the misalignment influences due to bounding box cropping. But there are obvious differences between these two strategies towards alignment issues. For RMGL, the ABP strategy is employed without additional training process according to the metric of activation balancing, which is not strongly correlated to the data distribution. For PCB+RPP, the RPP strategy learns the semantic consistency on each stripe partition and refines the part representation via the linear combination of prediction scores, which is still restricted by the alignment distribution.

\subsection{Random shifting augmentation}
Misalignment issues in stripe-based feature learning for person re-ID are mainly caused by three aspects: 1) captured views; 2) figure variances; 3) detection failures. Among all these difficulties, the first problem can be relieved by generative methods \cite{liu2018pose,qian2018pose} for more diverse view distribution or the guidance from pose estimation \cite{suh2018part}. In this section, we mainly focus on the latter two problems, both of which we find that the essence is that the person locating positions in misaligned pedestrian images are too different from the training distribution. So we introduce the random shifting augmentation (RSA) into person re-ID.

\begin{figure}
	\includegraphics[width=1.0\linewidth]{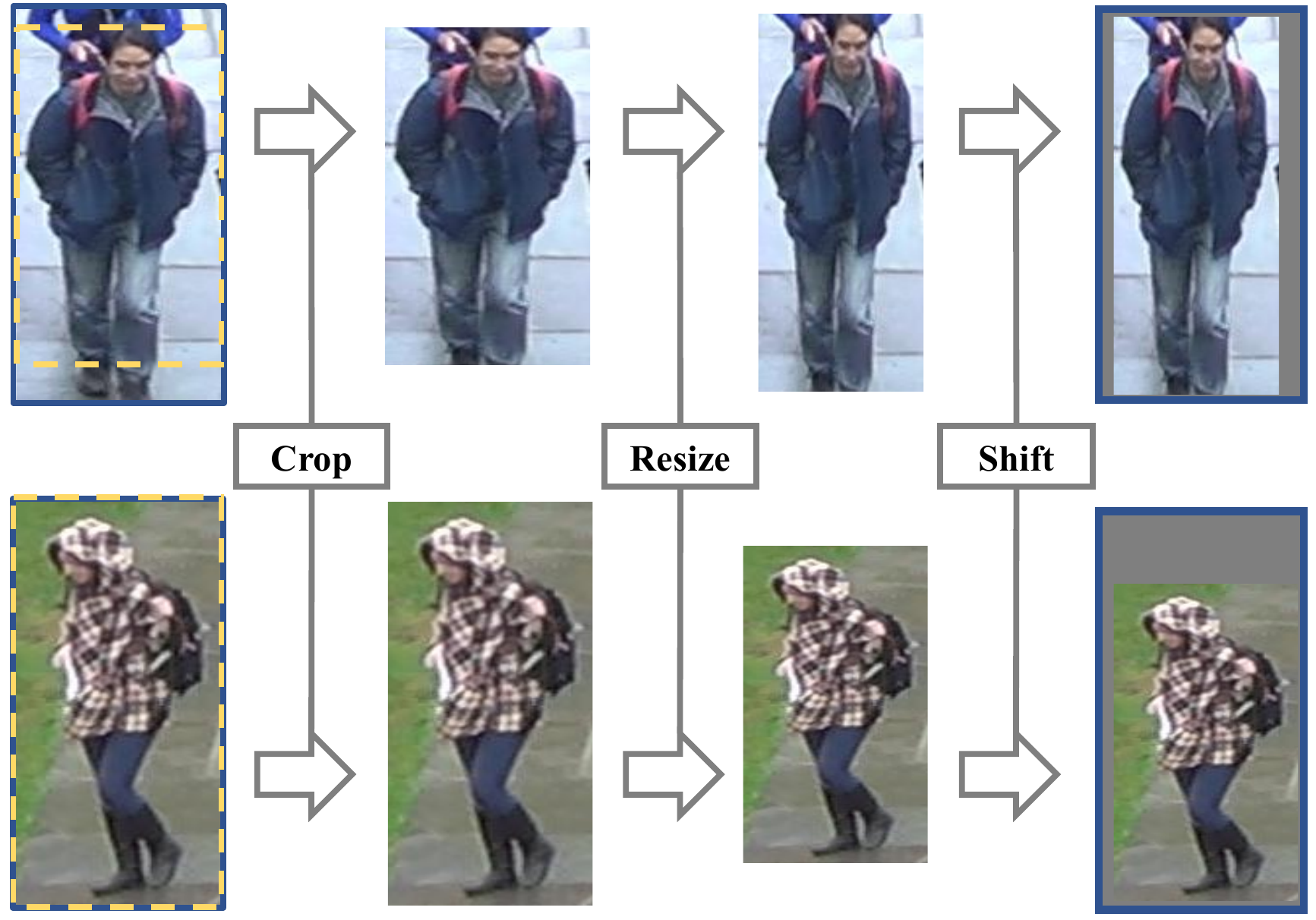}
	\centering
	\caption{\label{fig6}The pipeline of random shifting augmentation. The original image is randomly cropped to a smaller region or not. The yellow bounding boxes on both input images in ``Crop'' stage represent the random cropping regions. Here the image below is not randomly cropped. Then it is randomly resized by changing the aspect ratio. After that, the major person region is randomly located within a white box with average filling. }
\end{figure}

As illustrated in Fig. \ref{fig6}, the height and width of the original images are denoted as $H$ and $W$. The probability of randomly employing RSA is denoted as $p$. First, a randomly sampled area is cropped from an input pedestrian image with probability $p_c$. The crop ratio $r_c$ of cropped height to $H$ is subject to uniform distribution $U(r_c^{(min)}, 1.0)$. Then, the cropped area is re-scaled by a random rate $(r_h,r_w)$ respectively subject to independent uniform distributions $U(r_h^{(min)}, \frac{1}{r_c})$ and $U(r_w^{(min)}, 1.0)$. Finally, this resized patch can be shifted to a random location within a bounding box with the same size of original input images and ensure the patch will not locate outside this box. The uncovered area in the boxes is filled with constant values of 0.5 for float (or 127 for uint8) in all the RGB channels to simulate the average case of various backgrounds.

The effects of RSA come from the comprehensive efforts from the pipeline. The Crop step mainly augments the samples with missing body parts caused by detection failures. The Resize step achieves the random distortion in the dependency on human body figures in similarity metrics. The Shift step augments the cases of over-large bounding boxes predicted from inaccurate detection. The cooperation between Crop and Locate steps can also produce equivalent positive effects as Random Erasing \cite{zhong2017random} in particular cases.

\section{Experiments}
\subsection{Experimental settings}
\myPara{Datasets.~} All the experiments are conducted on four candidate datasets: Market-1501, DukeMTMC-reID, CUHK03 and MSMT17. Market-1501 includes training set with 12,936 images of 751 persons and testing set with 3,368 query and 19,732 gallery images of 750 persons. DukeMTMC-reID includes 16,522 images of 702 persons as the training set, and 2,228 query and 17,661 gallery images of 702 persons as the testing set. CUHK03 consists of 13,164 images with 1,360 persons from 6 cameras. MSMT17 consists of 126,441 images of 4,101 persons from 15 cameras. The cumulative matching characteristics (CMC) at rank-1, and mean average precision (mAP) are evaluated in both intra- and cross-dataset settings.

\myPara{Implementation.~} Input images are resized to $384 \times 128$ following \cite{sun2018beyond}. ResNet-50 is used as backbone network initialized with ImageNet \cite{deng2009imagenet} pretrained weights, two branches in RMGL are both initialized with the weights after $res\_conv2c$ resBlock in the original backbone. On input images, random horizontal flipping is employed with probability 0.5 besides the random shifting augmentations during training. The parameters of uniform distributions in RSA are set to $p_c=0.5$, $r_c=0.7$, $r_h^{(min)}=r_w^{(min)}=0.5$. For the batch hard triplet loss function, the margin $m$ is set to 0.6, and randomly sampling 12 persons with 4 images per person from the training set to construct image triplets. We choose SGD optimizer with momentum 0.9 to train our model. For the learning rate strategy, we first employ exponential warmup from 1e-5 to 0.01 in the starting 3 epochs to ease the hard process of multiple triplet losses optimization in early steps, and then employ decaying with factor 0.1 after training for 30 and 50 epochs. The weight decay for L2 regularization is set to 1e-3. For the network parts of 1x1 conv for reduction and fully connected layers for classification tasks, learning rates employed are 10 times the global rate. In evaluation, features from a pedestrian image and its horizontally flipped version are extracted, with the mean feature of both as the final feature. In our experiments, we use 2 NVIDIA TITAN Xp GPUs (12GB graphic memory usage) for data paralleling training and take about 3 hours for training on Market-1501 training set. In the inference time, it takes 20ms on average in testing.

\begin{table}[t]
	\caption{\label{marketduke}Intra-dataset comparison with other state-of-the-art methods on Market-1501 and DukeMTMC-reID datasets. Bold values represent the best results. }
	\centering
	\footnotesize
	\begin{tabular}{c|p{1.0cm}<{\centering}|p{1.0cm}<{\centering}|p{1.0cm}<{\centering}|p{1.0cm}<{\centering}}
		\hline
		\multirow{2}{*}{Methods} & \multicolumn{2}{c|}{Market-1501(SQ)} & \multicolumn{2}{c}{DukeMTMC-reID} \\
		\cline{2-5}
		& mAP & Rank-1 & mAP & Rank-1 \\
		\hline
		HA-CNN\cite{li2018harmoniou} & 75.7 & 91.2 & 63.8 & 80.5 \\
		DNN-CRF\cite{luo2018spectral} & 81.6 & 93.5 & 69.5 & 84.9 \\
		PCB+RPP\cite{sun2018beyond} & 81.6 & 93.8 & 69.2 & 83.3 \\
		Mancs\cite{wang2018mancs} & 82.3 & 92.3 & 71.8 & 84.9 \\
		GSRW\cite{shen2018deep} & 82.5 & 92.7 & 66.4 & 80.7 \\
		AANet-50\cite{tay2019aanet} & 82.5 & 93.9 & 72.6 & 86.4 \\
		SFT\cite{luo2018spectral} & 82.7 & 93.4 & 73.2 & 86.9 \\
		SGGNN\cite{shen2018person} & 82.8 & 92.3 & 68.2 & 81.1 \\
		CASN\cite{zheng2019ident} & 82.8 & 94.4 & 73.7 & 87.7 \\
		IANet\cite{hou2019interaction} & 83.1 & 94.4 & 73.4 & 87.1 \\
		HPM+HRE\cite{fu2019horizontal} & 83.1 & 93.9 & 74.5 & 86.3 \\
		CAMA\cite{yang2019towards} & 84.5 & 94.7 & 72.9 & 85.8 \\
		Auto-ReID\cite{quan2019auto} & 85.1 & 94.5 & - & - \\
		PAP+PS\cite{huang2018eanet} & 85.6 & 94.6 & 74.6 & 87.5 \\
		DG-Net\cite{zheng2019joint} & 86.0 & 94.8 & 74.8 & 86.6 \\
		MGN\cite{wang2018mgn} & 86.9 & 95.7 & 78.4 & 88.7 \\
		DSA-reID\cite{zhang2018densely} & 87.6 & 95.7 & 74.3 & 86.2 \\
		Pyramid\cite{zheng2018coarse} & 88.2 & 95.7 & 79.0 & 89.0 \\
		\hline
		RMGL(Ours) & \textbf{90.0} & \textbf{96.2} & \textbf{81.0} & \textbf{90.5} \\
		RMGL+RK & 95.6 & 96.7 & 91.0 & 92.6 \\
		\hline
	\end{tabular}
\end{table}

\begin{table}
	\caption{\label{cuhk03}Intra-dataset comparison with other state-of-the-art methods on Labeled and Detected images from CUHK03. }
	\centering
	\footnotesize
	\begin{tabular}{c|p{1.0cm}<{\centering}|p{1.0cm}<{\centering}|p{1.0cm}<{\centering}|p{1.0cm}<{\centering}}
		\hline
		\multirow{2}{*}{Methods}& \multicolumn{2}{c|}{Labeled} & \multicolumn{2}{c}{Detected} \\
		\cline{2-5}
		& mAP & Rank-1 & mAP & Rank-1  \\
		\hline
		HA-CNN\cite{li2018harmoniou}& 41.0 & 44.4 & 38.6 & 41.7 \\
		MLFN\cite{chang2018multilevel} & 49.2 & 54.7 & 47.8 & 52.8 \\
		PCB+RPP\cite{sun2018beyond} & - & -  & 57.5 & 63.7 \\
		SFT\cite{luo2018spectral} & 62.4 & 68.2 & - & - \\
		Mancs\cite{wang2018mancs} & 63.9 & 69.0 & 60.5 & 65.5 \\
		PAP+PS\cite{huang2018eanet} & - & - & 66.8 & 72.5 \\
		CAMA\cite{yang2019towards} & 66.5 & 70.1 & 64.2 & 66.6 \\
		CASN\cite{zheng2019ident} & 68.0 & 73.7 & 64.4 & 71.5 \\
		Auto-ReID\cite{quan2019auto} & 73.0 & 77.9 & 69.3 & 73.3 \\
		DSA-reID\cite{zhang2018densely} & 75.2 & 78.9 & 73.1 & 78.2 \\
		Pyramid\cite{zheng2018coarse} & 76.9 & 78.9 & 74.8 & 78.9 \\
		\hline
		RMGL(Ours) & \textbf{77.9} & \textbf{79.9} & \textbf{76.2} & \textbf{79.6}  \\
		RMGL+RK & 89.1 & 87.8 & 88.3 & 86.8 \\
		\hline
	\end{tabular}
\end{table}

\begin{table}
	\caption{\label{msmt17}Intra-dataset comparison with other state-of-the-art methods on MSMT17. }
	\centering
\footnotesize
	\begin{tabular}{c|p{1.0cm}<{\centering}|p{1.0cm}<{\centering}}
		\hline
		Methods & mAP & Rank-1 \\
		\hline
		GoogLeNet\cite{wei2018person} & 23.0 & 47.6\\
		PDC\cite{wei2018person} & 29.7 & 58.0 \\
		GLAD\cite{wei2018person} & 34.0 & 61.4 \\
		SFT\cite{luo2018spectral} & 47.6 & 73.6 \\
		IANet\cite{hou2019interaction} & 46.8 & 75.5 \\
		Auto-ReID\cite{quan2019auto} & 52.5 & 78.2 \\
		PAP+PS\cite{huang2018eanet} & 55.3 & 80.8 \\
		DG-Net\cite{zheng2019joint} & 52.3 & 77.2 \\
		\hline
		RMGL(Ours) & \textbf{60.4} & \textbf{83.6} \\
		\hline
	\end{tabular}
\end{table}

\subsection{Performance for Intra-Dataset Person re-ID}
We first compare our RMGL method with other methods for regular intra-dataset person re-ID. Table \ref{marketduke} shows the results on Market-1501 and DukeMTMC-reID, and Table \ref{cuhk03} and Table \ref{msmt17} respectively show the results on CUHK03 and MSMT17. Compared with Pyramid \cite{zheng2018coarse} based on ResNet-101 backbone, our method based on ResNet-50 backbone exceeds by 1.8\%@mAP/0.5\%@Rank-1 on Market-1501, 2.0\%@mAP/1.5\%@Rank-1 on DukeMTMC-reID, 1.0\%@mAP/1.0\%@Rank-1 on CUHK03 Labeled set and 1.4\%@mAP/0.7\%@Rank-1 on CUHK03 Detected set, proving that the performance advantages are not solely benefited from extra weights. Compared with MGN \cite{wang2018mgn} with trident multi-granularity learning architecture, our bident network surpasses by 3.1\%@mAP/0.5\%@Rank-1 on Market-1501 and 2.6\%@mAP/1.8\%@Rank-1 on DukeMTMC-reID, which shows that our proposed components comprehensively improve the representation efficiency with fewer branches. We establish a new state-of-the-art performance on MSMT17 benchmark with 60.4\%@mAP/83.6\%@Rank-1, exceeding all the earlier results by a large margin. Besides, RMGL surpasses all the re-ID methods guided by external cues (\textit{e.g.} PAP+PS\cite{huang2018eanet} and DSA-reID\cite{zhang2018densely}), which reveals the power of enhanced multi-granularity learning. After re-ranking post-processing \cite{zhong2017rerank}, the accuracy performances are dramatically improved by 5.6\%@mAP/0.5\%@Rank-1 on Market-1501, 10.0\%/mAP/2.1\%@Rank-1 and 11.2\%@mAP/9.9\%@Rank-1 on CUHK03-Labeled.

\begin{table}[t]
	\caption{\label{cross}Cross-dataset comparison with other state-of-the-art methods. The results are evaluated from the source dataset domains respectively from Market-1501, DukeMTMC-reID and CUHK03, to target dataset domains from Market-1501, DukeMTMC-reID, CUHK03, and MSMT17. The intra-domain results are not reported. \dag: Direct transferring results. }
	\footnotesize
	\centering
	\begin{tabular}{p{0.9cm}<{\centering}|p{1.52cm}<{\centering}|p{0.5cm}<{\centering}|p{0.5cm}<{\centering}|p{0.5cm}<{\centering}|p{0.5cm}<{\centering}|p{0.5cm}<{\centering}|p{0.5cm}<{\centering}}
		\hline
		\multirow{2}{*}{Source} & \multirow{2}{*}{Methods} & \multicolumn{2}{c|}{Market(SQ)} & \multicolumn{2}{c|}{Duke} & \multicolumn{2}{c}{CUHK03}  \\
		\cline{3-8}
		& & mAP & R1 & mAP & R1 & mAP & R1  \\
		\hline
		\multirow{6}{*}{Market} & IDE\cite{fan2018unsupervised}\dag & - & - & 10.9 & 21.9 & 4.4 & 5.5  \\
		& PUL\cite{fan2018unsupervised} & - & - & 16.4 & 30.4 & 7.3 & 7.6   \\
		& SPGAN\cite{deng2018image} & - & - & 22.3 & 41.1 & -  & -   \\
		& TJ-AIDL\cite{wang2018transferable} & - & - & 23.0 & 44.3  & -  & -  \\
		& HHL\cite{zhong2018generalizing} & - & - & 27.2 & 46.9 & -  & -    \\
		& PAP+PS\cite{huang2018eanet}\dag & - & - & 31.7 & 51.4  & 12.8  & 14.2 \\
		& RMGL(Ours)\dag & - & - & \textbf{33.8} & \textbf{52.4} & \textbf{13.6} & \textbf{14.8} \\
		\hline
		\multirow{6}{*}{Duke} & IDE\cite{fan2018unsupervised}\dag & 36.1 & 14.2 & - & - & 4.0 & 4.4  \\
		& PUL\cite{fan2018unsupervised} & 20.1 & 44.7 & - & - & 5.2 & 5.6   \\
		& SPGAN\cite{deng2018image} & 22.8 & 51.5 & - & - & -  & -   \\
		& TJ-AIDL\cite{wang2018transferable} & 26.5 & 58.2 & - & - & -  & -   \\
		& PAP+PS\cite{huang2018eanet}\dag & 32.9 & 61.7 & - & - & 9.6  & \textbf{11.4}   \\
		& HHL\cite{zhong2018generalizing} & 31.4 & 62.2 & - & - & -  & -  \\
		& RMGL(Ours)\dag & \textbf{33.4} & \textbf{63.7} & - & - & \textbf{9.9} & 11.0 \\
		\hline
		\multirow{5}{*}{CUHK03} & IDE\cite{fan2018unsupervised}\dag & 30.0 & 11.5 & 14.9 & 7.0 & -  & -    \\
		& PUL\cite{fan2018unsupervised} & 18.0 & 41.9 & 12.0 & 23.0  & -  & - \\
		& HHL\cite{zhong2018generalizing} & 29.8 & 56.8 & 23.4 & \textbf{42.7} & -  & - \\
		& PAP+PS\cite{huang2018eanet}\dag & 33.3 & 59.4 & 22.0 & 39.3& -  & -  \\
		& RMGL(Ours)\dag & \textbf{36.5} & \textbf{62.9} & \textbf{25.5} & 41.8& -  & -  \\
		\hline
		\multirow{2}{*}{MSMT17} & PAP+PS\cite{huang2018eanet}\dag & 37.9 & 66.4 & 46.4 & 67.4 & 17.4  & 19.4 \\
		& RMGL(Ours)\dag & \textbf{38.1} & \textbf{66.8} & \textbf{47.8} & \textbf{68.2} & \textbf{18.7}  & \textbf{20.8} \\
		\hline
	\end{tabular}
\end{table}

\subsection{Performance for Cross-Dataset Person re-ID}
In most previous contributions on person re-ID, the discrimination performance is mainly focused on accuracy figures in intra-dataset settings. However, this cannot adequately show the generalization in the true sense. Due to the inherent biases caused by sharing capture conditions, the learned representations easily suffer from overfitting to dataset biases by environmental factors. Here we complement the performance evaluation by cross-dataset person re-ID settings,  demonstrated in Table \ref{cross}. On the one hand, we compare RMGL with state-of-the-art unsupervised domain adaptation methods for person re-ID, by which the domain gap is bridged from unlabeled target domain samples. Surprisingly, our method actually outperforms all the listed UDA methods by a large margin. Compared with HHL \cite{zhong2018generalizing}, out method surpasses by 2.0\%@mAP/1.5\%@Rank-1. This result can reflect that even if no target domain samples can be utilized, RMGL can achieve pretty great generalization in representations for unknown dataset domains. On the other hand, compared with other supervised methods directly tranferred to  target domain, such as PAP+PS \cite{huang2018eanet}, RMGL can still outperform in cross-dataset settings, exceeding by 2.1\%@mAP/1.0\%@Rank-1 on Market-1501 to DukeMTMC-reID, and 2.0\%@mAP/0.5\%@Rank-1 on DukeMTMC-reID to Market-1501. For very-large-scale dataset MSMT17, our RMGL method can exceed with more advantages by 0.2\%@mAP/0.4\%@Rank-1, 1.4\%@mAP/0.8\%@Rank-1 and 1.3\%@mAP/1.4\%@Rank-1 respectively to Market-1501, DukeMTMC-reID and CUHK03. This result reflects that RMGL can achieve better generalization in most cases than methods guided by external cues.

\begin{table}
	\footnotesize
	\caption{\label{ablation}Ablation evaluation results by single- and multi-branch networks towards certain components. The model with $n$ split local partitions is denoted as $n$-stripe network, and $m$+$n$-stripe denotes the multi-granularity model. Here 2+3-stripe+RP+ABP+RSA is the same setting as our proposed RMGL model. BL: Baseline model.}
	\centering
	\label{tab:comparison_duke}
	\begin{tabular}{c|p{0.9cm}<{\centering}|p{0.9cm}<{\centering}|p{0.9cm}<{\centering}|p{0.9cm}<{\centering}}
		\hline
		\multirow{2}{*}{Methods} & \multicolumn{2}{c|}{Market-1501(SQ)} & \multicolumn{2}{c}{DukeMTMC-reID}\\
		\cline{2-5}
		& mAP & Rank-1 & mAP & Rank-1 \\
		\hline
		2-stripe (BL) & 85.6 & 95.0 & 76.5 & 88.1 \\
		3-stripe (BL) & 85.5 & 94.9 & 77.1 & 88.2 \\
		2+3-stripe (BL) & 87.1 & 95.5 & 79.0 & 89.3\\
		\hline
		2-stripe+RP & 86.8 & 95.0 & 77.5 & 88.1 \\
		3-stripe+RP & 86.6 & 95.3 & 78.3 & 88.9 \\
		2+3-stripe+RP & 88.3 & 95.7 & 79.9 & 89.5 \\
		2-stripe only RP & 86.3 & 95.0 & 77.2 & 88.0 \\
		3-stripe only RP & 86.5 & 95.3 & 78.1 & 88.7 \\
		2+3-stripe only RP & 88.0 & 95.7 & 79.6 & 89.3 \\
		\hline
		2-stripe+RSA & 86.8 & 95.2 & 77.4 & 88.9 \\
		3-stripe+RSA & 86.6 & 95.2 & 78.1 & 88.6 \\
		2+3-stripe+RSA & 88.3 & 95.9 & 79.9 & 89.5  \\
		2-stripe+RP+RSA & 88.1 & 95.5 & 78.7 & 89.1 \\
		2+3-stripe+RP+RSA & 89.7 & 96.1 & 80.8 & 90.3 \\
		\hline
		2-stripe+ABP & 86.0 & 95.2 & 76.7 & 88.5 \\
		3-stripe+ABP & 86.5 & 95.2 & 77.4 & 88.6 \\
		2+3-stripe+ABP & 87.8 & 95.5 & 79.3 & 89.7 \\
		2-stripe+RP+ABP+RSA & 88.5 & 95.6 & 79.0 & 89.1 \\
		2+3-stripe+RP+ABP+RSA & 90.0 & 96.2 & 81.0 & 90.5 \\
		\hline
		RMGL (1+2+3) & 89.8 & 96.0 & 81.2 & 90.5 \\
		RMGL (2+3+4) & 90.1 & 96.2 & 81.5 & 90.7 \\
		\hline
	\end{tabular}
\end{table}

\subsection{Ablation Study}
In this section, to verify the effectiveness of our proposed method, we conduct several ablation studies on the multi-granularity learning scheme, and components of receptive partition (RP), activation balanced pooling (ABP) and random shifting augmentation (RSA). The results in settings with different components are shown in Table \ref{ablation}. For baseline models, some settings are directly inherited from our proposed RMGL, which might be better than commonly used settings: 1) Parameter tuning: the learning rate decay is conducted only after 40 epoch in PCB setting, but after both 30 and 50 epoch in our baseline; the weight decay is set to 5e-4 in PCB setting, but 1e-3 in our baseline.  2) Global feature: global+stripe-local features are more powerful than pure stripe-local features according to \cite{wang2018mgn}, so we maintain these representation methods in both models. 3) Loss function: we maintain the triplet loss in our baseline.  Essentially, the reported baseline results in Table V are only used for comparison to verify our proposed modules. 

\myPara{Effectiveness of multi-granularity learning.~}
Comparing 2-stripe and 3-stripe models to 2+3-stripe models with different component settings, regardless of whether or not our proposed modules are applied, multi-branch networks with diverse granularities achieve performance that exceeds those single granularity networks. Among all the baseline networks, the multi-branch network respectively exceeds the best performance from single branch networks by 1.5\%@mAP/0.5\%@Rank-1 on Market-1501 and 1.9\%@mAP/1.1\%@Rank-1 on DukeMTMC-reID. These results show that representations with different levels of granularities from the multi-branch network indeed complement each other for better performance.

\myPara{Effectiveness of receptive partition.~}
We insert receptive partitioned paths achieved by RP after the $res\_conv2c$ layer in the backbone network. Compared with single branch baseline networks without receptive partitioned paths, 2-stripe with RP surpasses by 1.2\%@mAP on Market-1501 and 1.0\%@mAP on DukeMTMC-reID, and 3-stripe with RP surpasses by 1.1\%@mAP on Market-1501 and 1.2\%@mAP on DukeMTMC-reID. For the multi-branch networks, RP inserted in both branches can also bring considerable improvement by 1.2\%@mAP/0.2\%@Rank-1 on Market-1501 and 0.9\%@mAP/0.2\%@Rank-1 on DukeMTMC-reID. Compared with baseline models, the consistent improvement by RP accompanied models proves its positive effects on discrimination for feature learning. Besides, to validate the claim of priority of RP than OP, here we conduct several experiments with only RP, without the original path. Consistent improvement can also be achieved although slightly weak than dual-path settings, which verifies the remaining effects of the original path.

\myPara{Impact of partition numbers.~}
Partition numbers of RP determine the area of restricted receptive regions. Fig. \ref{split} illustrates the relationship between accuracy performance and partition numbers in stripe-based networks with RP on Market-1501 and DukeMTMC-reID datasets in single branch settings. The best partition setting changes with the dataset domains. For Market-1501 and CUHK03 datasets, the best stripe partition number is 2, and accuracy decreases along with the numbers increase, which might mean the over-fine-grained information in stripe-based learning might not be helpful. For DukeMTMC-reID and MSMT17 dataset, the best stripe partition number is 3 instead. We think this result might be related to the alignment biases of datasets.

\myPara{Impact of partition depth in network.~} 
The relationships between Rank-1/mAP and hyper-parameters in the single branch backbone network are illustrated in Fig. \ref{dd}. For the cases of K = 2 and 3, the best results are achieved when RP is employed on the intermediate feature map instead of Input Partition or Output Partition, and the RP result is still comparable to the best result of OP for K = 6, reflecting the necessity of RP. Meanwhile, results of K = 2 and 3 obviously outperform K = 6 in all conditions, which might be caused by insufficient receptive region for feature maps split into more K stripes. For the case of K = 6, an overall upward trend for precision before splitting at res conv3 reveals the fact of the swelling region from larger receptive field sizes, and the degradation cliff of Rank-1 between res3 and res4 might be caused by the influences of redundancy.

\begin{figure}
	\centering
	\subfigure[$p_c$]{
		\centering
		\includegraphics[width=0.297\linewidth]{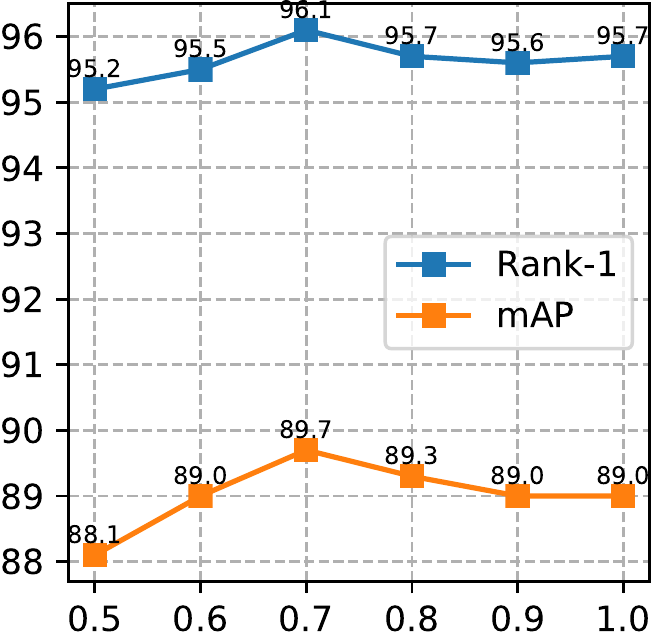}
	}
	\subfigure[$r_c$]{
		\centering
		\includegraphics[width=0.297\linewidth]{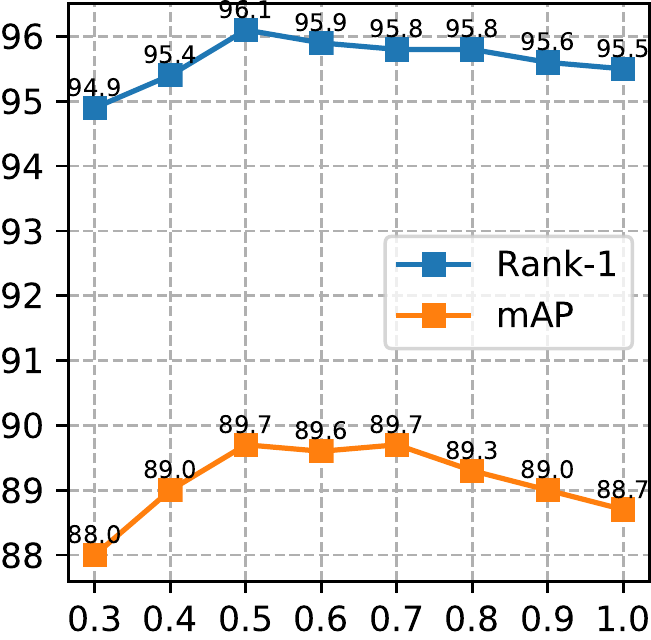}
	}
	\subfigure[$r_w^{(min)}$]{
		\centering
		\includegraphics[width=0.297\linewidth]{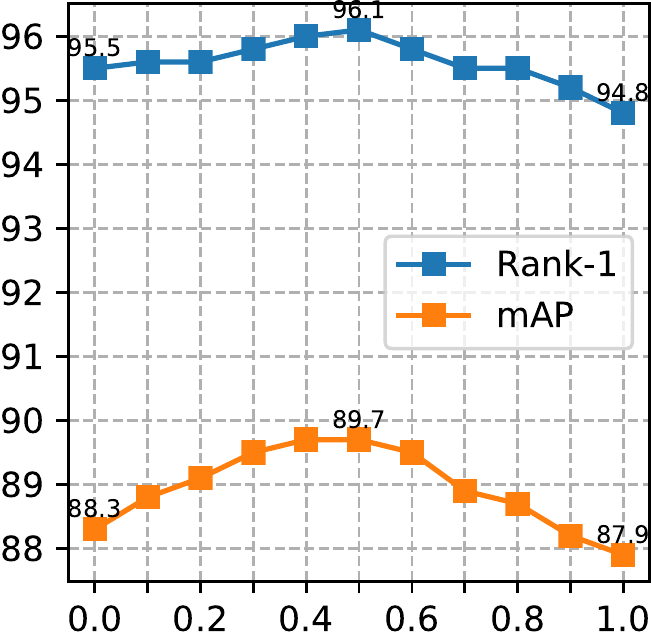}
	}
	\caption{\label{rs} Impact of random shifting parameters $p_c$, $r_c$ and $r_w^{(min)}$ evaluated on the Market-1501 dataset. For (b) and (c), some extreme values are not evaluated for its obvious degradation. All the hyper-parameters are evaluated with other parameters fixed. }
\end{figure}

\begin{table}
	\footnotesize
	\caption{\label{aug} \footnotesize Effects by data augmentations commonly used in person re-ID. RCJ: Random Color Jittering. RC: Random Cropping. RE: Random Erasing. RSA: random shifting augmentation.}
	\centering
	\begin{tabular}{c|c|c|c|c}
		\hline
		\multirow{2}{*}{Methods} & \multicolumn{2}{c|}{Market} & \multicolumn{2}{c}{Duke}\\
		\cline{2-5}
		& mAP & Rank-1 & mAP & Rank-1 \\
		\hline
		2-stripe+RCJ & 85.8$_{+0.2}$ & 94.9$_{-0.1}$& 76.4$_{-0.1}$& 88.1$_{0.0}$  \\
		2-stripe+RC & 85.6$_{0.0}$ & 94.8$_{-0.2}$ & 76.6$_{+0.1}$ & 88.0$_{-0.1}$ \\
		2-stripe+RE & 86.4$_{+0.8}$ & 95.0$_{0.0}$ & 77.1$_{+0.6}$ & 88.5$_{+0.4}$  \\
		2-stripe+RCRS & 86.0$_{+0.4}$ & 95.1$_{+0.1}$  & 76.8$_{+0.3}$ & 88.0$_{-0.1}$  \\		
		2-stripe+RSA & 86.8$_{+1.2}$ & 95.2$_{+0.2}$  & 77.4$_{+0.9}$ & 88.9$_{+0.8}$  \\		
		2-stripe+RSA+RE& \textbf{87.3}$_\mathbf{+1.7}$ & \textbf{95.4}$_\mathbf{+0.4}$ & \textbf{77.8}$_\mathbf{+1.3}$ & \textbf{89.3}$_\mathbf{+1.2}$  \\
		\hline
		2+3-stripe+RCJ & 87.2$_{+0.1}$  & 95.3$_{-0.2}$ & 79.2$_{+0.2}$ & 89.3$_{0.0}$ \\
		2+3-stripe+RC & 87.3$_{+0.2}$ & 95.3$_{-0.2}$  & 79.3$_{+0.3}$ & 89.2$_{-0.1}$ \\
		2+3-stripe+RE & 88.0$_{+0.9}$ & 95.7$_{+0.2}$ & 79.5$_{+0.5}$ & 89.5$_{+0.2}$ \\		
		2+3-stripe+RCRS & 87.7$_{+0.6}$ & 95.6$_{+0.1}$  & 79.4$_{+0.4}$ & 89.3$_{0.0}$  \\		
		2+3-stripe+RSA & {88.3}$_{+1.2}$ & {95.9}$_{+0.4}$ & {79.9}$_{+0.9}$ & 89.5$_{+0.2}$ \\	
		2+3-stripe+RSA+RE & \textbf{89.0}$_\mathbf{+1.9}$ & \textbf{96.1}$_\mathbf{+0.6}$  & \textbf{80.2}$_\mathbf{+1.2}$ & \textbf{89.5}$_\mathbf{+0.2}$  \\
		\hline
	\end{tabular}
\end{table}

\myPara{Effectiveness of activation balanced pooling.~}
On Market-1501, compared with baseline models, ABP brings mAP improvement by 0.4\% for the 2-stripe network, 1.0\% for 3-stripe network, and 0.7\% for the multi-branch network. Compared with networks powered by RP and RSA, ABP brings mAP improvement by 0.4\% for the 2-stripe network and 0.3\% for the multi-branch network. On DukeMTMC-reID, compared with baseline models, ABP brings mAP improvement by 0.2\% for the 2-stripe network, 0.3\% for 3-stripe network and 0.3\% for multi-branch network, and with RSA the improvements are  0.3\% and 0.2\% respectively on the 2-stripe single branch and multi-branch networks. Although we can observe both improvement effects on both benchmarks, the effectiveness is more obvious on the Market-1501. Considering that Market-1501 is misaligned worse than DukeMTMC-reID, this reveals that ABP is more effective in more hard-aligned conditions. Surely the benefits from ABP are not as significant as RP or RSA, but the performance improvement on both datasets verify that ABP provides us another potential direction to determine better coarse part areas instead of uniform partitions. We can also find the effect of ABP is weaker when being combined with the RP and RSA. This result is mainly caused by random shifting augmentation. RSA aims at distorting the inherent distribution of pedestrian body placing to enhance the robustness against spatial misalignment. This distortion method also increases a stripe variance for the activation significance balancing metric, which might influence the stability of ABP.

\myPara{Effectiveness of random shifting augmentation.~}
RSA brings more significant improvement on models with various settings. On 2-stripe baseline networks, accuracy improvement are achieved respectively by 1.2\%@mAP/0.2\%@Rank-1 and 0.9\%@mAP/0.8\%@Rank-1 on Market-1501 and DukeMTMC-reID. 3-stripe networks with RSA are improved by 1.1\%@mAP/0.3\%@Rank-1 and 1.0\%@mAP/0.4\%@Rank-1 on both datasets. For multi-branch settings, 1.2\%@mAP/0.4\%@Rank-1 and 0.9\%@mAP/0.2\%@Rank-1 improvement on both datasets can also be achieved. On model settings applied with RP, 2-stripe settings with RSA surpasses by 2.3\%@mAP/0.5\%@Rank-1 and 1.2\%@mAP/1.0\%@Rank-1 on both datasets, while 2+3-stripe surpasses by 1.4\%@mAP/0.4\%@Rank-1 and 0.9\%@mAP/0.7\%@Rank-1. On the one hand, the larger accuracy gain by RSA+RP might be brought by larger local area variance from RSA, enhanced the generalized locality by RP. On the other hand, the weaker improvement on multi-branch networks probably means that locality diversity might have brought similar cross-covering effects by RSA.

\begin{figure}
	\centering
	\subfigure[Market-1501]{
		\centering
		\includegraphics[width=0.46\linewidth]{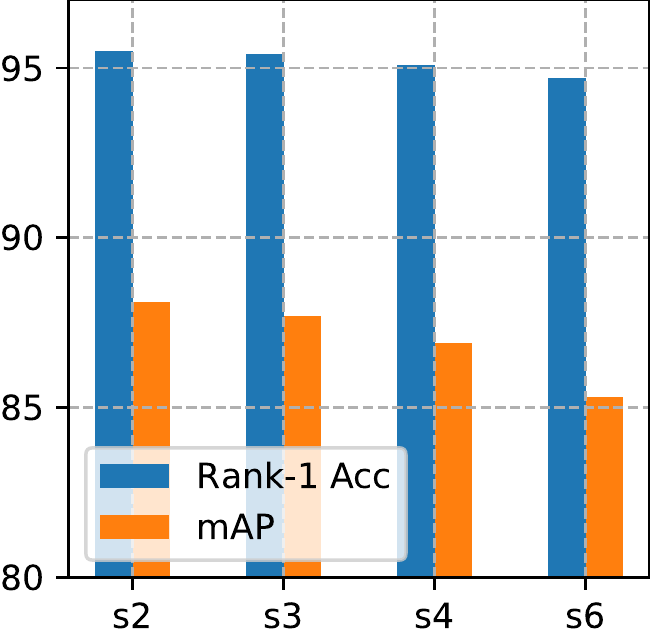}
	}
	\subfigure[DukeMTMC-reID]{
		\centering
		\includegraphics[width=0.46\linewidth]{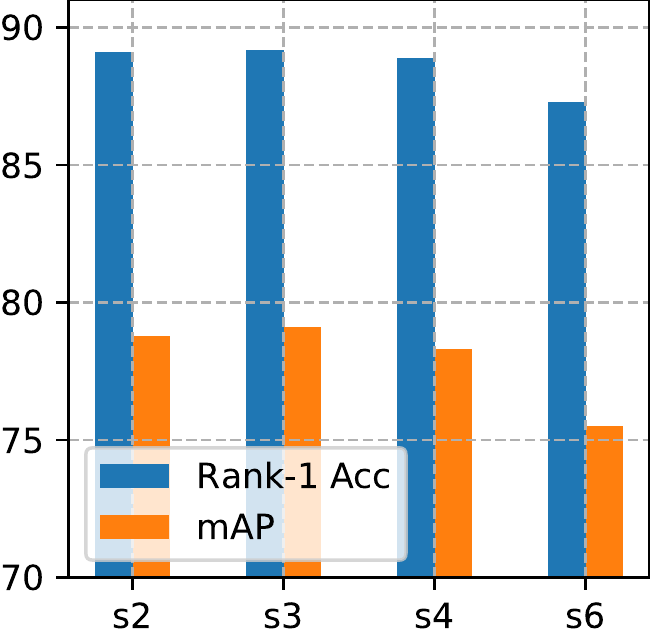}
	}
	\subfigure[CUHK03]{
	\centering
	\includegraphics[width=0.46\linewidth]{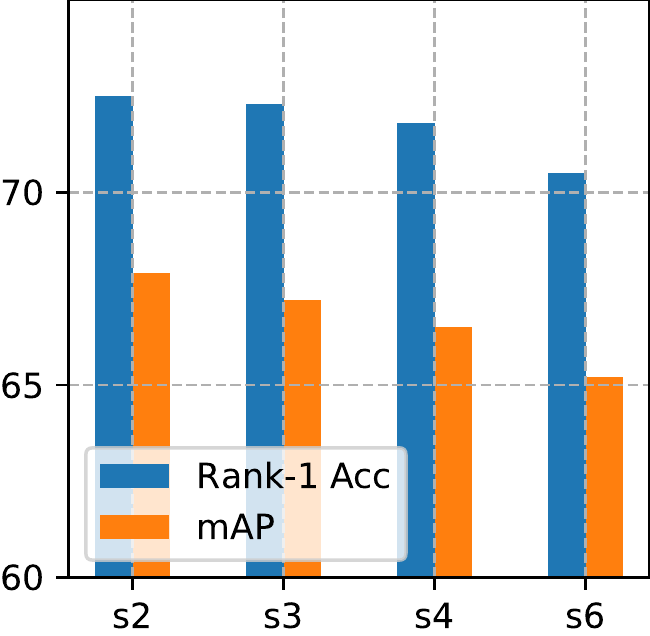}
	}
	\subfigure[MSMT17]{
	\centering
	\includegraphics[width=0.46\linewidth]{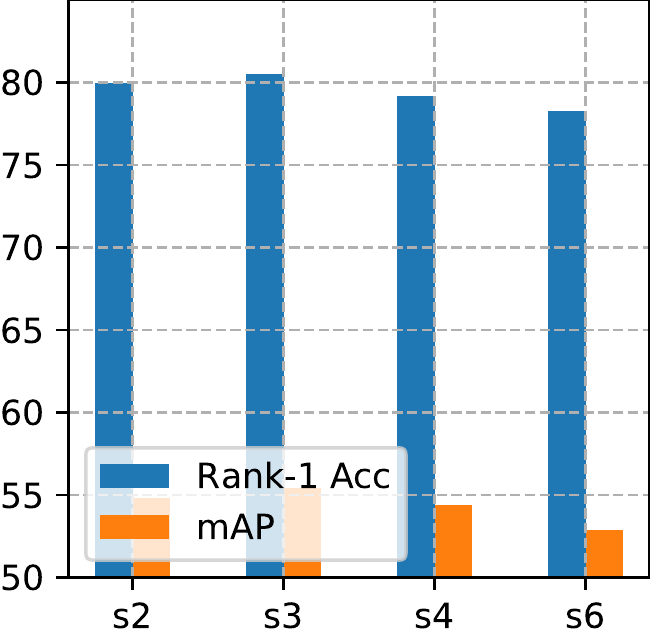}
	}	
	\caption{\label{split}Relationship between Rank-1 accuracy/mAP and partition numbers, evaluated on Market-1501, DukeMTMC-reID, CUHK03, and MSMT17 datasets, evaluated by single-branch dual-path networks with RP and RSA. ``s\_$n$" refers to the partition number of $n$.}
\end{figure}

\myPara{RSA vs other data augmentations.~}
We compare with major augmentations in re-ID such as random color jittering \cite{yang2019path}, cropping, and erasing \cite{zhong2017random}. Notice that we've used flipping by default. No GAN-based and auto-augmentation methods are compared to avoid effects by external data and unaffordable computing costs. Augmentations are all tuned to their best settings.Table \ref{aug} shows results by methods above on our baseline models.  Some standard augmentations seem not to work well: proper random color jittering extends the reflecting tendency for some colors, which has no explicit help on the performance intra-domain person re-ID for the fixed capturing condition; random cropping introduces biases of person cross-cutting edges, which seems not to bring any positive effects on feature learning. Random erasing is the most effective augmentation method except for RSA, which generates samples with occlusion. The similar performance improvement of RSA shows that distraction on placing the distribution of human bodies on captured person images also has significant effects on feature learning just as occlusion. We also conduct experiments with both RE and RSA augmentations. This indeed brings further improvement based on model with only RSA, but the accuracy gain by RE is a bit weaker than that on without RSA, implying that RSA might have effects on augmenting edge occluded pedestrian examples as RE. RSA is similar to a random ``crop, resize and shift'' pipeline (RCRS) that includes three steps: i)~cropping a proper random region with arbitrary sizes, ii)~resizing the cropped image region to a predefined input size, and iii)~applying a random translation coordinate transformation (\textit{i.e.} $T=[[1,0,t_x];[0,1,t_y]]$). From Table VI, we can find that ``RCRS'' does not bring similar accuracy improvement as RSA does. The ``shift-after-resize'' operation is applied on the exact box of resized images, and will definitely bring out-of-range information loss from translated images, which might dramatically shrink the effective input contents. Besides, direct resizing operation after cropping would have brought unsatisfactory aspect distortion. 

\begin{figure}
	\subfigure[Rank-1 accuracy]{
	\centering
	\includegraphics[width=0.46\linewidth]{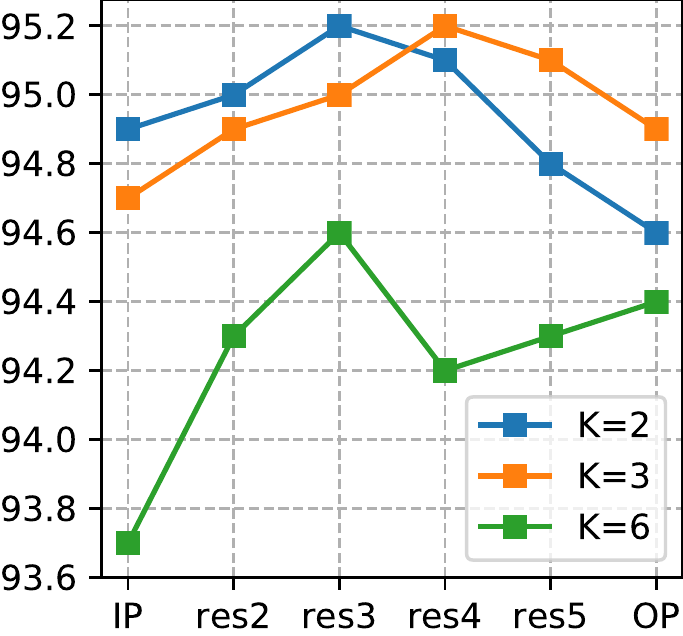}
	}
	\subfigure[mAP]{
	\centering
	\includegraphics[width=0.46\linewidth]{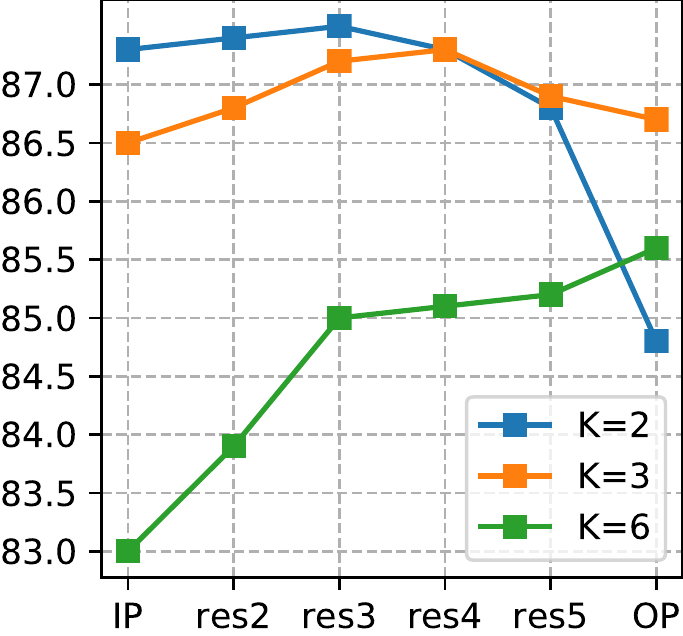}
	}
	\centering
	\caption{\label{dd}Relationship between Rank-1 accuracy/mAP and splitting positions/number of stripes, evaluated on Market-1501 dataset. Here \textit{res\_x} refers to the position where feature maps are fed into the resBlock $res\_conv\_x$. }
\end{figure}

\myPara{Analysis on the random shifting parameters.~} 
Random shifting augmentations involves a series of parameters. Fig. \ref{rs} shows the impact of $r_s$, $r_c$ and $r_w^{(min)}$ parameters. Notice that $r_h^{(min)}$ is not considered for its upper bound decided by $r_c$. For $p_c$, a nearly flatten performance range can be reached between the values from 0.3 to 0.6, with the best value 0.5. To a definite augmentation by $p_c=1$, degradation can be caused by over large placing variation. For $r_c$, the best parameter is set to 0.7. The accuracy seems to be worse when $r_c$ is smaller than around 0.5, but steady when $r_c$ is relatively large.  The accuracy is obviously degraded when the lower bound $r_w^{(min)}$, gets too small. Combining $r_c$ and $r_w^{(min)}$, we can find that when the cropping rate is set to be large, the effects by random shifting might be dramatically limited, which shows the necessity of placing randomly sampled partitions.

\myPara{Impact of the branch number in RMGL.~}
Comparing to the multi-granularity learning method \cite{wang2018mgn} , we only introduce bident network branches instead of a trident. But this does not mean that one more extra branch cannot bring further improvement. So we add another branch based on RMGL: model with an extra Local-4 Branch inserted by receptive partitioned path denoted as RMGL(2+3+4), and with an extra Global Branch denoted as RMGL(1+2+3) network. Compared with bident RMGL, both trident RMGL brings no significant improvement on Market-1501, and only a slight improvement on DukeMTMC-reID. But considering the trident RMGL with the receptive partitioned path is much more complicated and slower than the bident, and introduces limited performance improvement, we prefer the trade-off between performance and efficiency, so our proposed RMGL is still bident.

\begin{figure}
	\centering
	\subfigure[Market-1501]{
		\centering
		\includegraphics[width=0.46\linewidth]{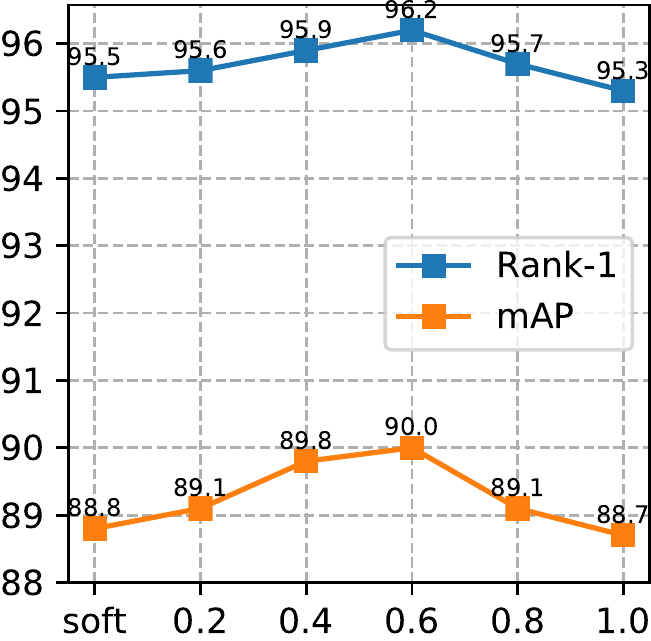}
	}
	\subfigure[DukeMTMC-reID]{
		\centering
		\includegraphics[width=0.46\linewidth]{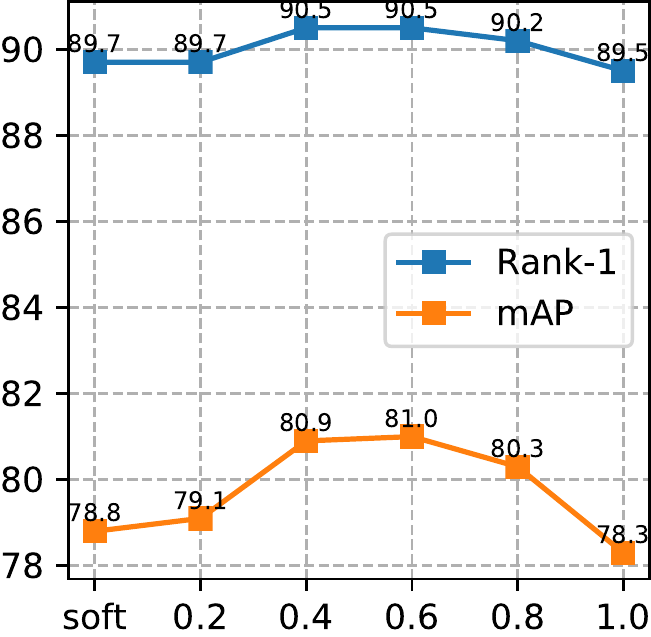}
	}
	\caption{\label{triplet}Relationship between Rank-1 accuracy/mAP and margin parameter for triplet losses, evaluated on Market-1501 and DukeMTMC-reID datasets. ``soft'' means the soft-margin triplet loss in \cite{hermans2017defense} without specific margin settings. }
\end{figure}

\myPara{Analysis on the triplet margin.~} 
Notice that as the multi-embedding concatenation with mutually independent triplet losses refinement, we can theoretically tune various margin parameters for different triplet losses. To ease the complexity, all the margins are unified to one single value. As shown in Fig. \ref{triplet}, we can find that with the margin around 0.4 to 0.6, the RMGL model can reach a better accuracy level on both datasets. An over large or small margin setting can relatively degrade the discrimination. Particularly, the soft-margin triplet loss \cite{hermans2017defense} is also considered in this analysis for hyper-parameter reduction. However, in our setting, it seems to perform worse than any setting with specific margins. So we use a specific margin of 0.6 in the implementation.

\begin{figure}
	\centering
	\subfigure[Rank-1 accuracy]{
		\centering
		\includegraphics[width=0.46\linewidth]{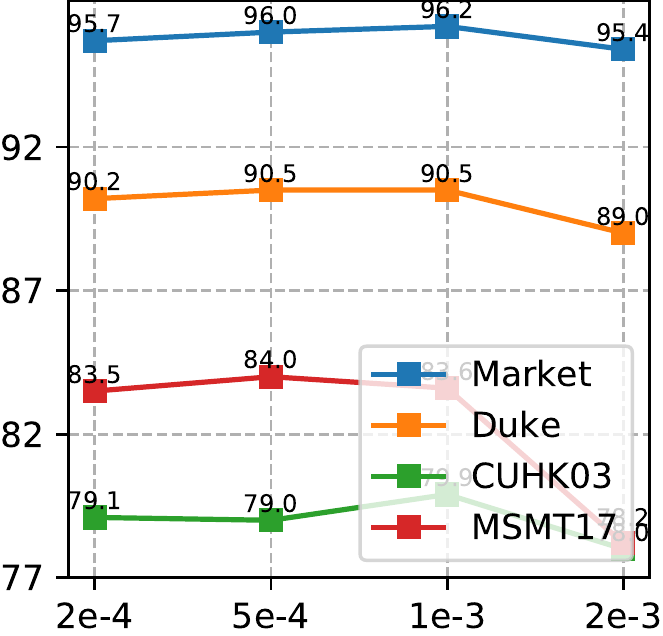}
	}
	\subfigure[mAP]{
		\centering
		\includegraphics[width=0.46\linewidth]{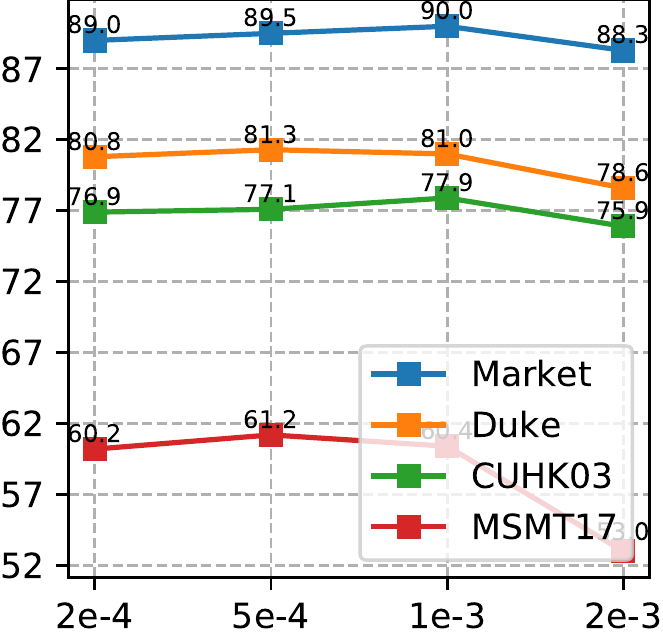}
	}
	\caption{\label{wd}Relationship between Rank-1 accuracy/mAP and weight decay factors, evaluated on Market-1501, DukeMTMC-reID, CUHK03(Labeled) and MSMT17 datasets. }
\end{figure}

\myPara{Analysis on weight decay.~} 
Due to the limited scale of re-ID dataset, regularization might have effects on the discrimination. Fig. \ref{wd} shows the impact of weight decay factors on the performance. On Market-1501 and CUHK03 datasets, 1e-3 is the best weight decay value, respectively exceeds 0.5\%@mAP/0.2\%@Rank-1 and 0.8\%@mAP/0.9\%@mAP in 5e-4 setting. On MSMT17, 1e-3 is worse than 5e-4 by 0.8\%@mAP/0.4\%@Rank-1. On DukeMTMC-reID, 1e-3 and 5e-4 settings share a similar performance level and even 5e-4 is better in mAP. 2e-4 is an acceptable setting for training on all datasets, but the 2e-3 factor dramatically degrades the performance. Combining all the situations of different datasets, we choose 1e-3 as the final setting.

\section{Conclusion}
In this paper, we propose the receptive multi-granularity learning (RMGL) algorithm to represent local identity features with enhanced body detail locality and diversity, achieved by our proposed novel multi-branch receptive partitioned deep network architecture. Furthermore, we design the activation balanced pooling as the splitting approach instead of uniform partitions, which introduce local representations by balanced semantic significance. We also introduce simple but effective random shifting augmentation to confront misalignment issues. Extensive comprehensive and ablation experiments with intra-dataset and cross-dataset scenarios prove the outstanding discrimination and generalization by our proposed architecture in person re-ID feature learning.

\myPara{Acknowledgement.~} This work was partially supported by grants from the National Natural Science Foundation of China (61772513), and the project from Beijing Municipal Science and Technology Commission (Z191100007119002). Shiming Ge is also supported by the Open Projects Program of National Laboratory of Pattern Recognition, and the Youth Innovation Promotion Association, Chinese Academy of Sciences.

{\small
	\bibliographystyle{IEEEtran}
	\bibliography{bibMGN}
}

\begin{IEEEbiography}[{\includegraphics[width=1in,height=1.25in,clip,keepaspectratio]{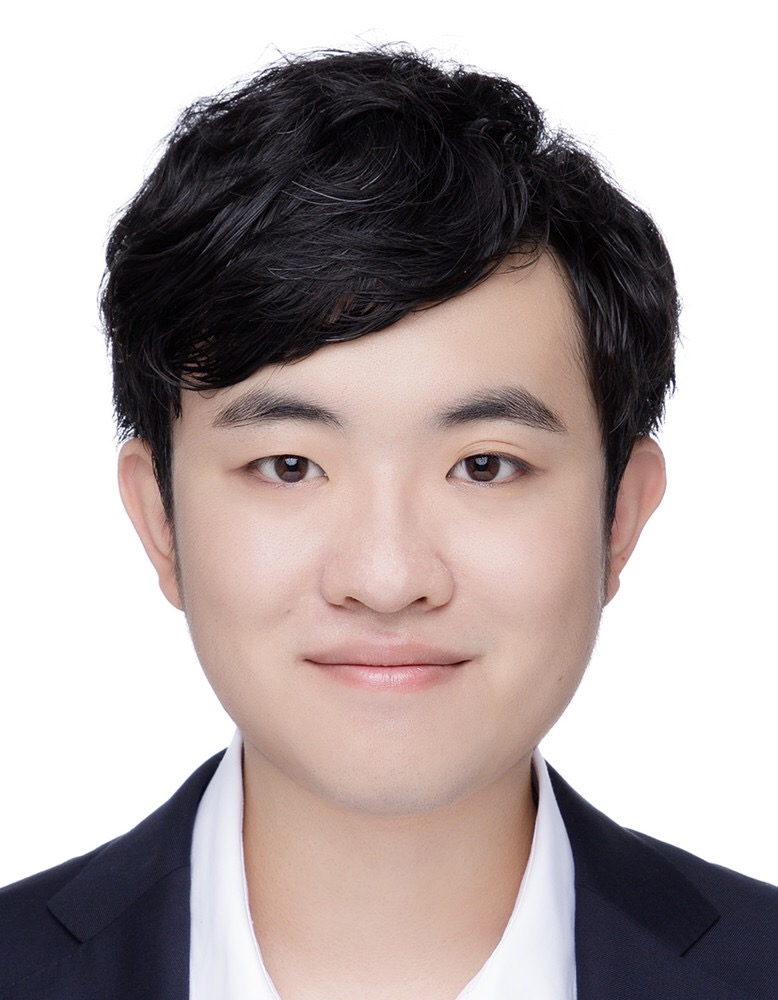}}]{Guanshuo Wang} is currently pursuing the Ph.D. degree in Shanghai Jiao Tong University, China. He received the B.S. degree in electronic engineering from Xidian University, China, in 2016. His research interests include person re-identification, face recognition and transfer learning.
\end{IEEEbiography}

\begin{IEEEbiography}[{\includegraphics[width=1in,height=1.25in,clip]{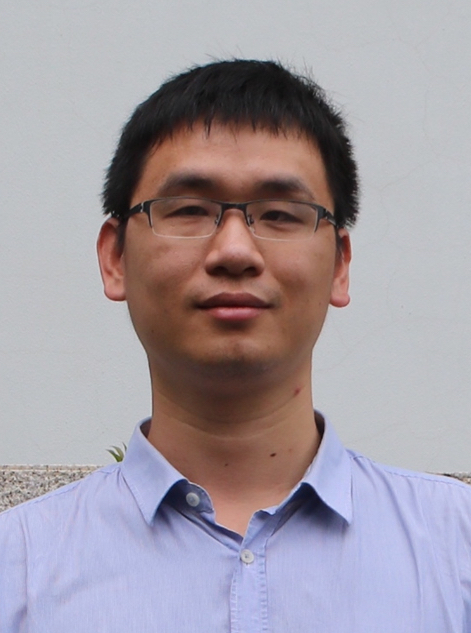}}]{Yufeng Yuan} is currently a computer vision researcher in CloudWalk Research, China.  He received the M.S. degree in Zhejiang University, China. His research interests include person re-identification, video analysis and automated machine learning.
\end{IEEEbiography}

\begin{IEEEbiography}[{\includegraphics[width=1in,height=1.25in,clip,keepaspectratio]{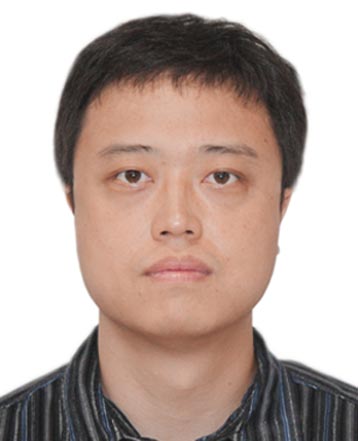}}]{Jiwei Li} is dean of CloudWalk Research, China. He received the B.S. and M.S. degrees from the University of Science and Technology of China. His research interests include computer vision, video analysis, speech recognition.
\end{IEEEbiography}

\begin{IEEEbiography}[{\includegraphics[width=1in,height=1.25in,clip,keepaspectratio]{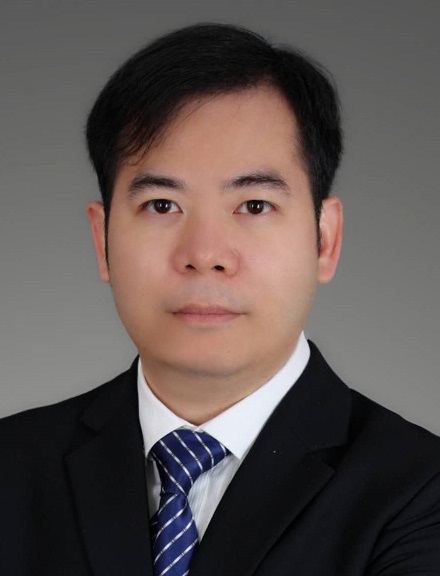}}]{Shiming Ge}
(M'13-SM'15) is currently an Associate Professor with the Institute of Information Engineering at Chinese Academy of Sciences. He is also the member of Youth Innovation Promotion Association, Chinese Academy of Sciences. Prior to that, he was a project manager in ShanDa Innovations, a researcher in Samsung Electronics and Nokia Research Center. He received the B.S. and Ph.D degrees both in Electronic Engineering from the University of Science and Technology of China (USTC) in 2003 and 2008, respectively. His research mainly focuses on computer vision, data analysis, machine learning and AI security, especially efficient learning models and solutions toward scalable applications.
\end{IEEEbiography}

\begin{IEEEbiography}[{\includegraphics[width=1in,height=1.25in,clip,keepaspectratio]{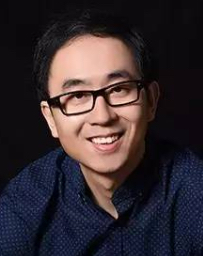}}]{Xi Zhou} is currently the president of CloudWalk Technology, China. He is also a specially appointed Professor and a Ph.D. advisor of Shanghai Jiao Tong University, China. Before that, he was a Professor of Chongqing Institute of Green and Intelligent Technology, Chinese Academy of Sciences. He received the B.S. and M.S. degrees from the University of Science and Technology of China, in 2003 and 2006, and received the Ph.D. degree from Department of Electrical and Computer Engineering of University of Illinois at Urbana-Champaign, in 2010. He has published more than 40 papers, and is the holder of more than 300 authorized patents. He has won 6 world-class computer vision contests, including the ImageNet Large Scale Visual Recognition Challenge (ILSVRC) in 2010.
\end{IEEEbiography}

\end{document}